\documentclass[10pt,twocolumn,letterpaper]{article}

\usepackage[pagenumbers]{iccv} % To force page numbers, e.g. for an arXiv version
\usepackage{pgfplots}
\usepackage{pgfplotstable}
\usepackage{siunitx}
\usepackage{graphicx}
\usepackage{graphicx}
\usepackage{booktabs}
\usepackage{multirow}
\usepackage{makecell}
\usepackage{array}
\usepackage{booktabs}
\usepackage{threeparttable}

\definecolor{iccvblue}{rgb}{0.21,0.49,0.74}
\usepackage[pagebackref,breaklinks,colorlinks,allcolors=iccvblue]{hyperref}

\def\paperID{12874} % *** Enter the Paper ID here
\def\confName{ICCV}
\def\confYear{2025}

\newcommand{\wliu}[1]{}

\newcommand{\model}{DiT-Air}

\title{\model: Revisiting the Efficiency of Diffusion Model Architecture Design \\ in Text to Image Generation}

\author{Chen Chen,\,\, Rui Qian,\,\, Wenze Hu,\,\, Tsu-Jui Fu,\,\,  Jialing Tong,\,\,  Xinze Wang,\,\, \\ Lezhi Li,\,\, Bowen Zhang,\,\, Alex Schwing,\,\, Wei Liu,\,\, Yinfei Yang\,\, \\
Apple Inc. \\ 
{\tt\small \{chen\_chen999\}@apple.com}
\vspace{0.3cm} % To keep exactly the same layout as with the REVIEW version.
}

\begin{document}
\maketitle
\begin{abstract}
In this work, we empirically study Diffusion Transformers (DiTs) for text-to-image generation, focusing on architectural choices, text-conditioning strategies, and training protocols. We evaluate a range of DiT-based architectures--including PixArt-style and MMDiT variants--and compare them with a standard DiT variant which directly processes concatenated text and noise inputs. 
Surprisingly, our findings reveal that the performance of standard DiT is comparable with those specialized models, while demonstrating superior parameter-efficiency, especially when scaled up. Leveraging the layer-wise parameter sharing strategy, we achieve a further reduction of 66\% in model size compared to an MMDiT architecture, with minimal performance impact. Building on an in-depth analysis of critical components such as text encoders and Variational Auto-Encoders (VAEs), we introduce {\model} and {\model-Lite}. With supervised and reward fine-tuning, {\model} achieves state-of-the-art performance on GenEval and T2I CompBench, while {\model-Lite} remains highly competitive, surpassing most existing models despite its compact size.
\end{abstract}    
\section{Introduction}

The field of text-to-image synthesis has witnessed remarkable progress, primarily attributable to the wide adoption of diffusion-based models~\cite{ho2020denoising, esser2024scaling}. Diffusion Transformers (DiTs)~\cite{peebles2023scalable} have emerged as a prominent architectural paradigm, combining the iterative denoising process inherent to diffusion models with the representational efficacy of transformer networks. While existing paradigm variants like PixArt-style ~\cite{chenpixart, chen2024pixartdeltafastcontrollableimage, chen2024pixartsigmaweaktostrongtrainingdiffusion} and MMDiT from Stable Diffusion 3~(SD3)~\cite{esser2024scaling} have demonstrated strong performance, key aspects of DiTs—such as the choice of architectural components, text-conditioning mechanisms, and training strategies—have not been exhaustively explored yet~\cite{hatamizadeh2024diffitdiffusionvisiontransformers,chen2024gentrondiffusiontransformersimage}.

% Link to graph below: https://docs.google.com/spreadsheets/d/1PhCq0NYVCS_E6M90-BU0gagub1pl2llJv1kAhkmFcd4/edit?gid=0#gid=0
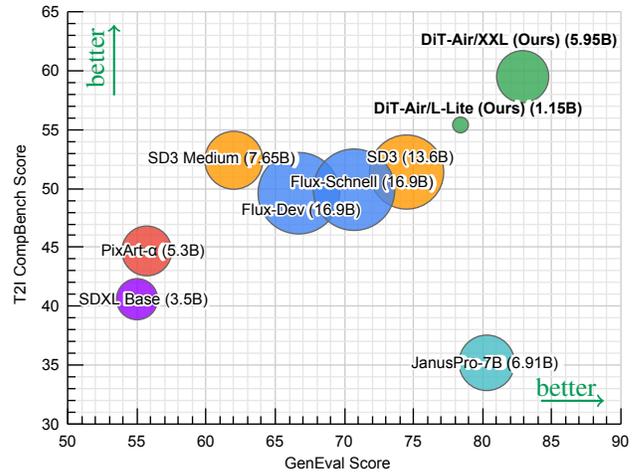
\begin{figure}
    \centering
    \begin{tikzpicture}
        % Include the image
        \node[anchor=south west, inner sep=0] (image) at (0,0) 
            {\includegraphics[width=1.0\linewidth, trim=30 35 10 20, clip]{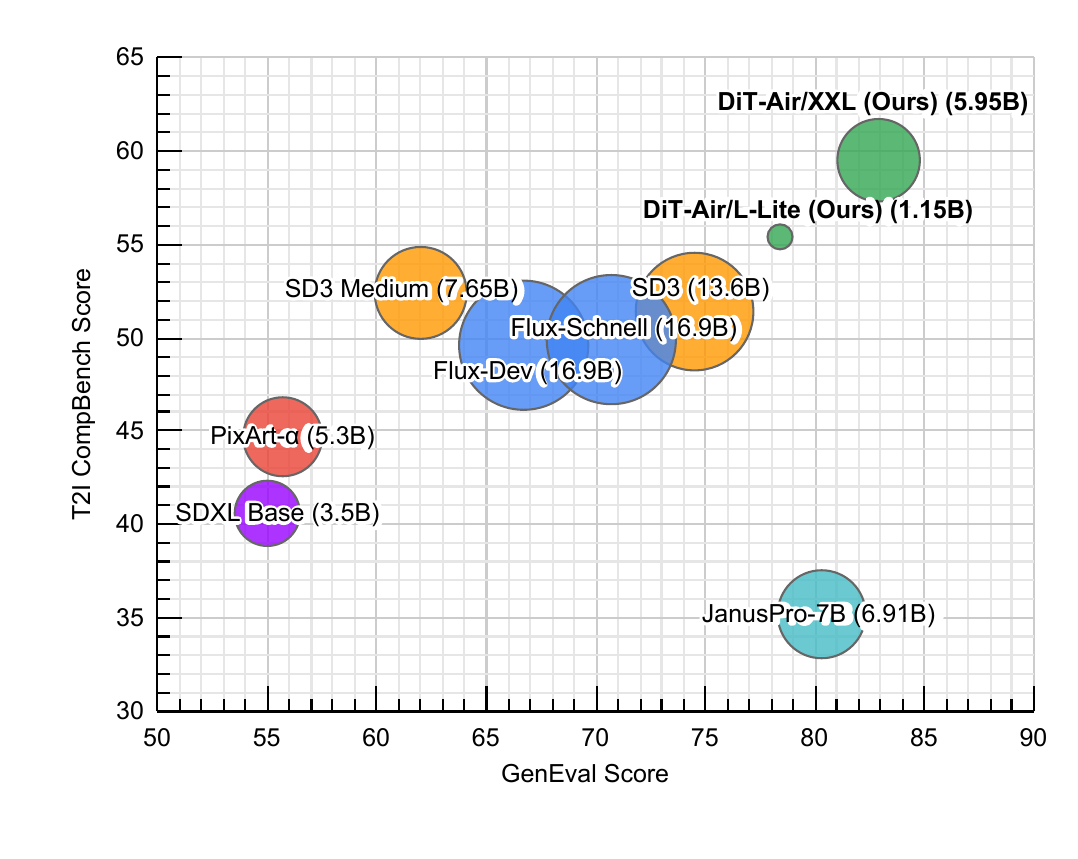}};

        % Get the width and height of the image
        \begin{scope}[shift={(image.south west)}, x={(image.south east)}, y={(image.north west)}]
            
            % Arrow pointing UP (better T2I CompBench Score)
            \draw[thick, ->, ForestGreen] (0.17, 0.8) -- (0.17, 0.95);
            \node[ForestGreen, rotate=90] at (0.14, 0.88) {{better}};

            % Arrow pointing RIGHT (better GenEval Score)
            \draw[thick, ->, ForestGreen] (0.85, 0.15) -- (0.95, 0.15);
            \node[ForestGreen] at (0.89, 0.18) {better};
        
        \end{scope}
    \end{tikzpicture}
    \caption{Comparison of text-to-image generation methods on two metrics, GenEval and T2I CompBench (higher is better for both). Despite significantly smaller model size, our proposed \model\ achieves state-of-the-art results. Note that, for our model, we report the full model size including text encoder and VAE. A detailed parameter breakdown is provided in Appendix~\ref{sec:appendix_sota_model_size_breakdown}.}
    \label{fig:teaser}
\end{figure}

In this work, we conduct a comprehensive investigation into DiT design choices for text-to-image synthesis. Beginning with a comparative analysis of the vanilla DiT~\cite{peebles2023scalable}, PixArt-$\alpha$~\cite{chenpixart}, and MMDiT~\cite{esser2024scaling}, we develop a streamlined architecture. This architecture utilizes concatenated text and noise inputs (following MMDiT) and shared AdaLN parameters (following PixArt-$\alpha$), eliminating modality-specific projections. This simplification yields substantial parameter savings (66\% compared to MMDiT and 25\% compared to PixArt-$\alpha$), while preserving or enhancing performance. 
\begin{figure*}[t]
    \centering
    \includegraphics[width=\textwidth, trim=110 125 110 110, clip]{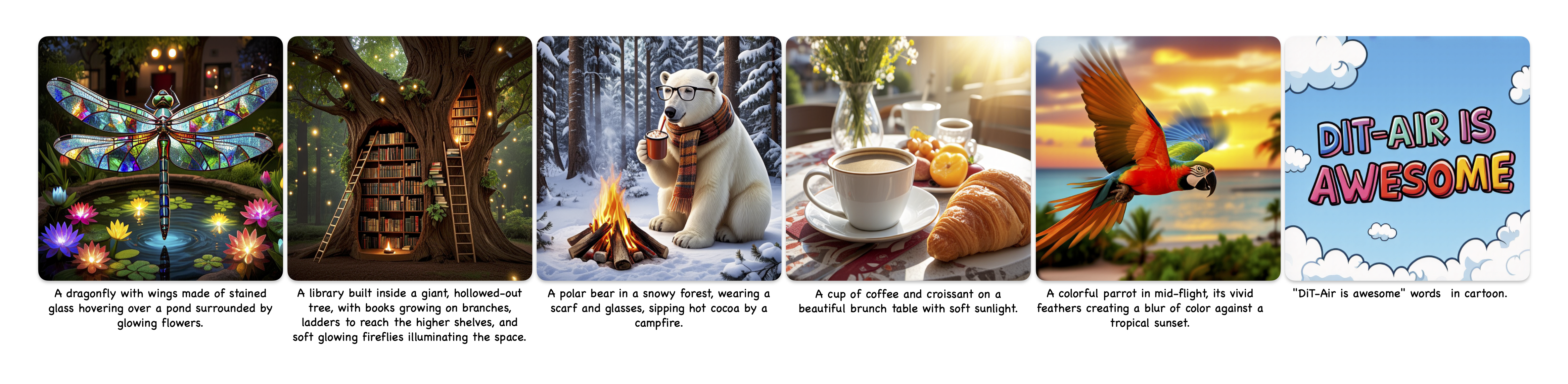}
    \caption{Sample images from our proposed \model, each with the text prompt below it. See Appendix~\ref{sec:appendix_samples} for more examples.}
    \label{fig:samples_banner}
\end{figure*}
Notably, the resulting architecture, named \model, closely resembles the original DiT,  allowing us to leverage existing transformer optimizations.
Inspired by parameter-sharing strategies in NLP models such as ALBERT~\cite{lanalbert}, we adopt both full block-sharing and attention-only sharing schemes to further push the parameter efficiency of DiT models. Our ablation studies demonstrate that attention-only sharing provides a compelling trade-off, achieving significant parameter reduction with minimal loss in text alignment and generative fidelity.

In addition to architectural innovations, \model\ benefits from a thorough analysis of text-conditioning strategies and variational autoencoders~(VAEs). Specifically, we evaluate three primary types of text encoders: CLIP, large language models (LLMs), and the T5 model. Our study includes comprehensive ablations of causal versus bidirectional CLIP, layer selection strategies for both CLIP and LLMs, and a final performance comparison of all three encoders. We also introduce a refined variational autoencoder (VAE)~\cite{kingma2022autoencodingvariationalbayes} that better preserves fine-grained visual details, further boosting image quality, especially complex visual features.

Finally, %upon building upon our previous findings and further 
with a progressive training approach, \model\ %, we present a family of models, also named {\bf \model} and {\bf \model-Lite}, 
%achieve 
achieves new state-of-the-art  GenEval~\cite{ghosh2023genevalobjectfocusedframeworkevaluating} and T2I CompBench~\cite{huang2025t2i} scores of 82.9 and 59.5, respectively. As shown in Figure~\ref{fig:teaser}, our model delivers superior performance with outstanding parameter efficiency compared to leading models such as SD3, FLUX, and JanusPro. Example generation results are provided in Figure~\ref{fig:samples_banner}.

Our key contributions are  as follows: 
($i$) We systematically study the design choices of a range of DiT-based architectures including PixArt-style and MMDiT variants.
($ii$) We introduce {\model} and \model-Lite, a novel DiT model family that simply extends a standard DiT by directly processing concatenated text and noise inputs. 
($iii$) We demonstrate parameter efficiency, achieving a 66\% model size reduction with minimal performance impact compared to the state-of-the-art MMDiT. 
($iv$) We establish a new state-of-the-art performance on GenEval~\cite{ghosh2023genevalobjectfocusedframeworkevaluating} and T2I CompBench~\cite{huang2025t2i}.

\section{Related Works}
\label{sec:related}

\subsection{Text-to-Image Diffusion Models}

Diffusion models~\cite{song2020score, ho2020denoising, dhariwal2021diffusion} have achieved remarkable success in text-to-image generation~\cite{ramesh2022hierarchicaltextconditionalimagegeneratio, saharia2022photorealistictexttoimagediffusionmodels}. These models generate images by iteratively denoising random noise, guided by semantic text embeddings extracted from pre-trained text encoders, such as CLIP~\cite{radford2021learningtransferablevisualmodels}. Latent diffusion methods~\cite{rombach2022high, podell2023sdxl} further enhanced training and inference efficiency by operating in a latent space defined by a pre-trained variational autoencoder (VAE)~\cite{kingma2022autoencodingvariationalbayes}, reducing computational costs without sacrificing quality. Recently, flow matching objectives~\cite{liu2022flow, ma2024sit, tong2023improving} have been introduced to connect source and target image distributions through simplified paths, offering further gains in image quality~\cite{esser2024scaling}. Our approach builds upon this paradigm by leveraging pre-trained text encoders for conditioning and optimizing a conditional flow matching objective in latent space.

\subsection{Diffusion Transformers and Text Conditioning}

Diffusion Transformers (DiTs)~\cite{peebles2023scalable} were initially proposed to extend the advantages of transformer architectures to class-conditional image generation within diffusion models. Built upon vanilla vision transformers~\cite{dosovitskiy2020image}, widely utilized in image understanding tasks, DiT explored various approaches for incorporating noise level (time) and class label conditions. The study compared zero-initialized adaptive layer normalization (AdaLN), cross-attention, and in-context conditioning, demonstrating that AdaLN was the most effective strategy through comprehensive experiments.

Subsequent works extended DiTs to text-to-image generation, establishing them as a popular choice for high-quality open-source models~\cite{chenpixart, esser2024scaling, flux2024}. Among them, PixArt models use cross attention between fixed text embeddings and image features after each of the self attention layers to inject text conditions into transformer models. MMDiT expands the transformer to a dual-stream design, with separate query, key, value, and output (QKVO) projections and MLPs for text embeddings and image features. Text and image features interact via scaled dot product attention applied on concatenated feature sequences. Compared to vanilla DiTs with a similar number of layers and feature dimensions, both PixArt and MMDiT expand the model size and use the extra parameters to convert semantically rich text embeddings into the visual space.

To justify these extra complications for text conditioning, the SD3 paper~\cite{esser2024scaling} reports that MMDiT outperforms several DiT variants. However, this study was conducted on the relatively small CC12M~\cite{changpinyo2021conceptual} dataset and at smaller model scales, without fully accounting for the additional parameters introduced by MMDiT's dual-stream architecture. Moreover, a scaling law study by Liang et al.~\cite{liang2024scaling} compared cross-attention with in-context conditioning but excluded AdaLN and focused only on models with 1M to 1B parameters, limiting its applicability to larger-scale settings. Through large-scale experiments, we demonstrate that our \model, which adheres closely to the simpler vanilla DiT architecture, can achieve comparable or superior performance to these more complex models, particularly at scale.

%Despite promising results, the two text conditioning approaches have not been systematically compared against each other through carefully designed large scale experiments. Furthermore, with the observation that there is minimal separate treatments of image and text tokens in large scale vision language models, it is unclear why significant amount of extra parameters are necessary to accommodate text information in diffusion models. 
%We show that the added complexities do not necessarily lead to better models, especially when models are scaled up and trained on large scale datasets. SD3 reports that MMDiT outperforms several DiT variants, its study uses the relatively small CC12M dataset and smaller model scales, without fully accounting for the additional parameters introduced by MMDiT’s two-stream architecture. Moreover,

\paragraph{Parameter Sharing in Transformers.}

Parameter sharing has emerged as an effective approach to enhance efficiency in transformer-based architectures. For instance, ALBERT~\cite{lanalbert} demonstrates that sharing parameters across layers in a BERT-like model can substantially reduce model size while retaining competitive performance on NLP tasks. The parameter sharing approach in our proposed diffusion model architecture (Section~\ref{sec:monodit-lite}) is inspired by this work, and offers new avenues for balancing generative quality and model compactness in DiT architectures.

%Inspired by this, our proposed \model-lite variants (Section~\ref{sec:\model-lite}) apply similar parameter-sharing concepts to text-to-image diffusion, offering new avenues for balancing generative quality and model compactness in DiT architectures.

\section{Architecture Design}
\label{sec:model arch}
%-------------------------------------------------------------------------
\subsection{Latent Diffusion Framework}
\label{sec:latent_diffusion}

A typical text-to-image generation pipeline based on latent diffusion encodes the target image \(\mathbf{x}\) into a latent representation \(\mathbf{z}_0 \in \mathbb{R}^{h \times w \times c}\) via a variational autoencoder (VAE), \ie, \(\mathbf{z}_0 = E_{\text{VAE}}(\mathbf{x})\). Meanwhile, the text prompt \(p\) is processed by one or more pretrained text encoders (\eg, CLIP, T5), and the resulting embeddings are projected to produce token embeddings \(\mathbf{c} \in \mathbb{R}^{l_\text{text} \times d}\).

\begin{figure}[t]
\begin{center}
\includegraphics[width=0.41\textwidth]{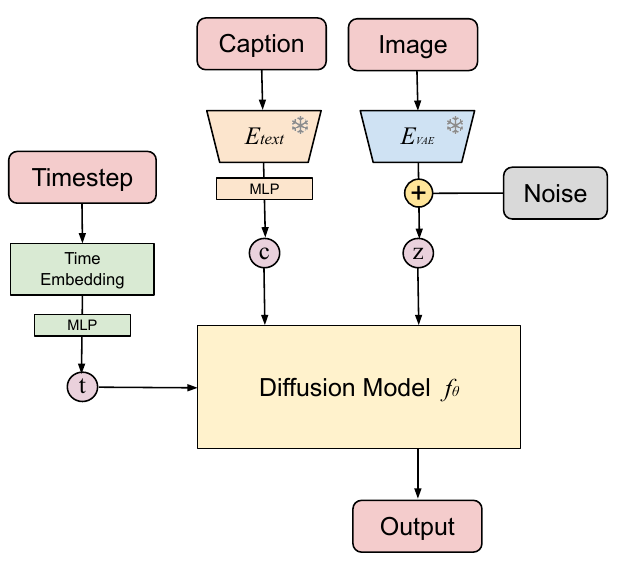}
\end{center}
\caption{\textbf{Overview of Latent Diffusion Training.} During training, \(\mathbf{x}\) is encoded into a latent \(\mathbf{z}_0\) via a VAE, and the text prompt \(p\) is mapped to embeddings \(\mathbf{c}\). A forward diffusion adds noise to \(\mathbf{z}_0\), and the model learns to reverse this process by predicting the noise (or similar target) at each timestep. \wliu{change E\_text to blue color as well, indicating it is frozen.}}
\label{fig:latent_diffusion}
\vspace{-0.42cm}
\end{figure}

During training (see Figure~\ref{fig:latent_diffusion}), a forward diffusion process adds noise to \(\mathbf{z}_0\), producing a noisy latent \(\mathbf{z}_t\) for a randomly sampled \(t \in (0, 1)\). The model \(f_\theta\) then learns to reverse this process by predicting a target quantity---such as noise \(\epsilon\) in denoising models, velocity \(\mathbf{v}\) in v-prediction models, or vector field \(\mathbf{F}\) in flow-matching models---conditioned on the current latent \(\mathbf{z}_t\), the text embedding \(\mathbf{c}\), and the timestep \(t\):
\[
\hat{y}_t = f_\theta(\mathbf{z}_t, \mathbf{c}, t).
\]

The training objective minimizes the discrepancy between \(\hat{y}_t\) and the true target, effectively denoising \(\mathbf{z}_t\). At inference, \(f_\theta\) iteratively transforms a random sample \(\mathbf{z}_T\) back to a clean latent \(\mathbf{\hat{z}}_0\), which is then decoded by the VAE to produce the final image.

%-------------------------------------------------------------------------

\subsection{Text-conditioned Diffusion Transformers}
\label{sec:dit_backbone}

In this section, we examine the backbone diffusion model \(f_\theta\), specifically focusing on two widely-used Diffusion Transformer (DiT) variants: PixArt-$\alpha$ and MMDiT. 

\begin{figure*}[t]
    \centering
    \newcommand{\imageheight}{7.5cm}
    \begin{tabular}{c@{\hskip -2mm}c@{\hskip -4mm}c} 
        \hspace{-10mm}\includegraphics[height=\imageheight]{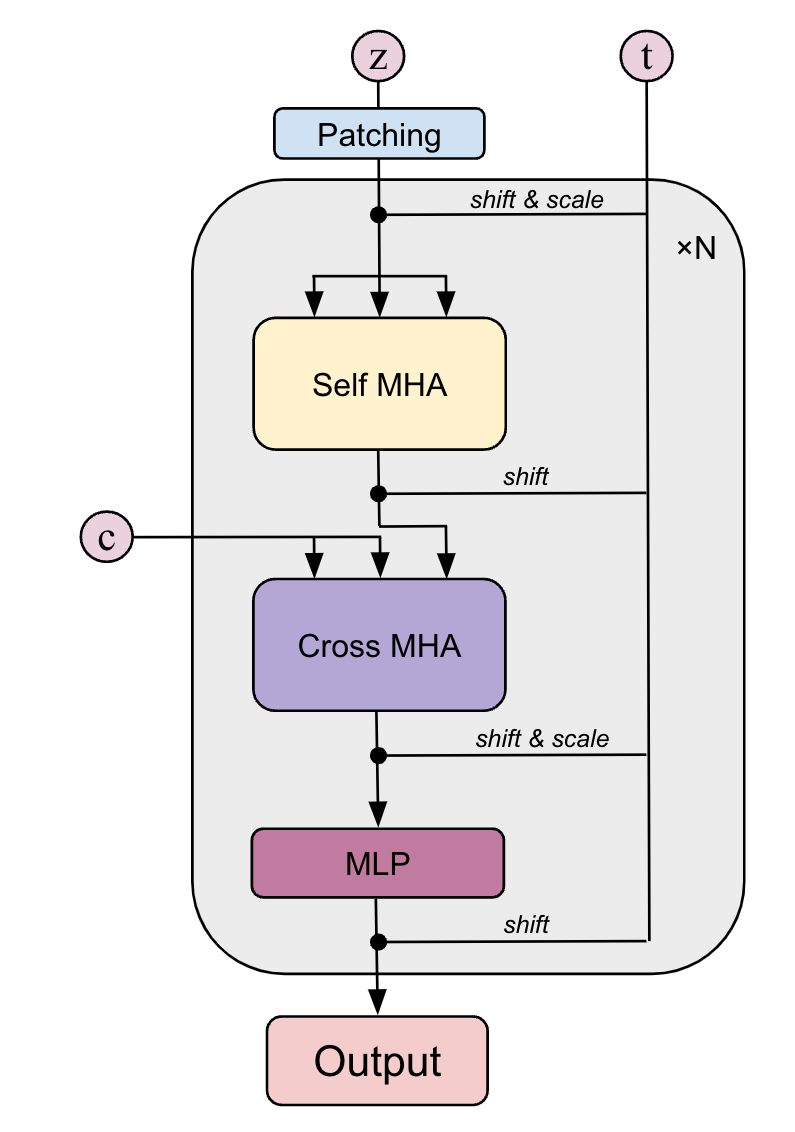} &
        \includegraphics[height=\imageheight]{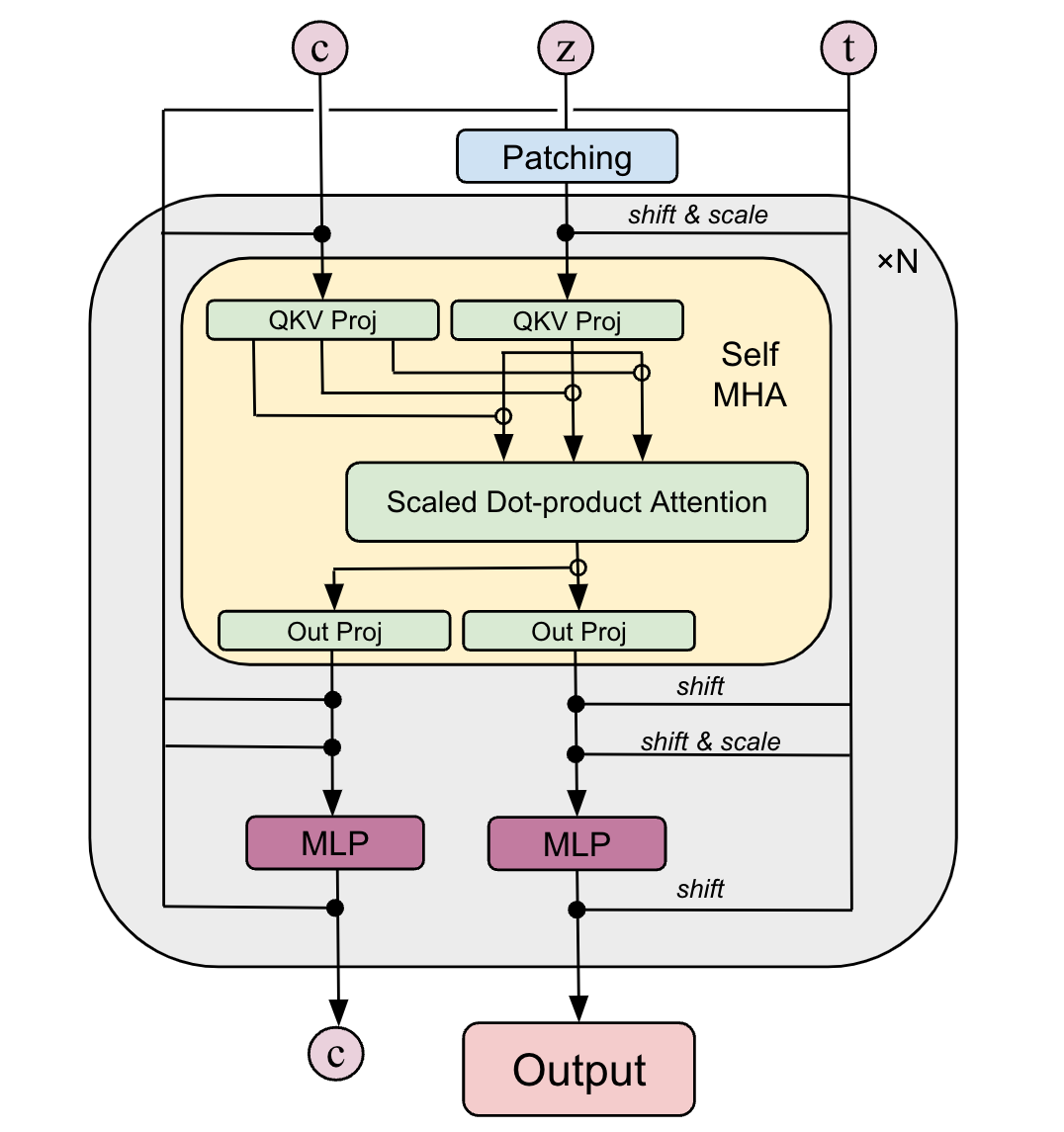} &
        \includegraphics[height=\imageheight]{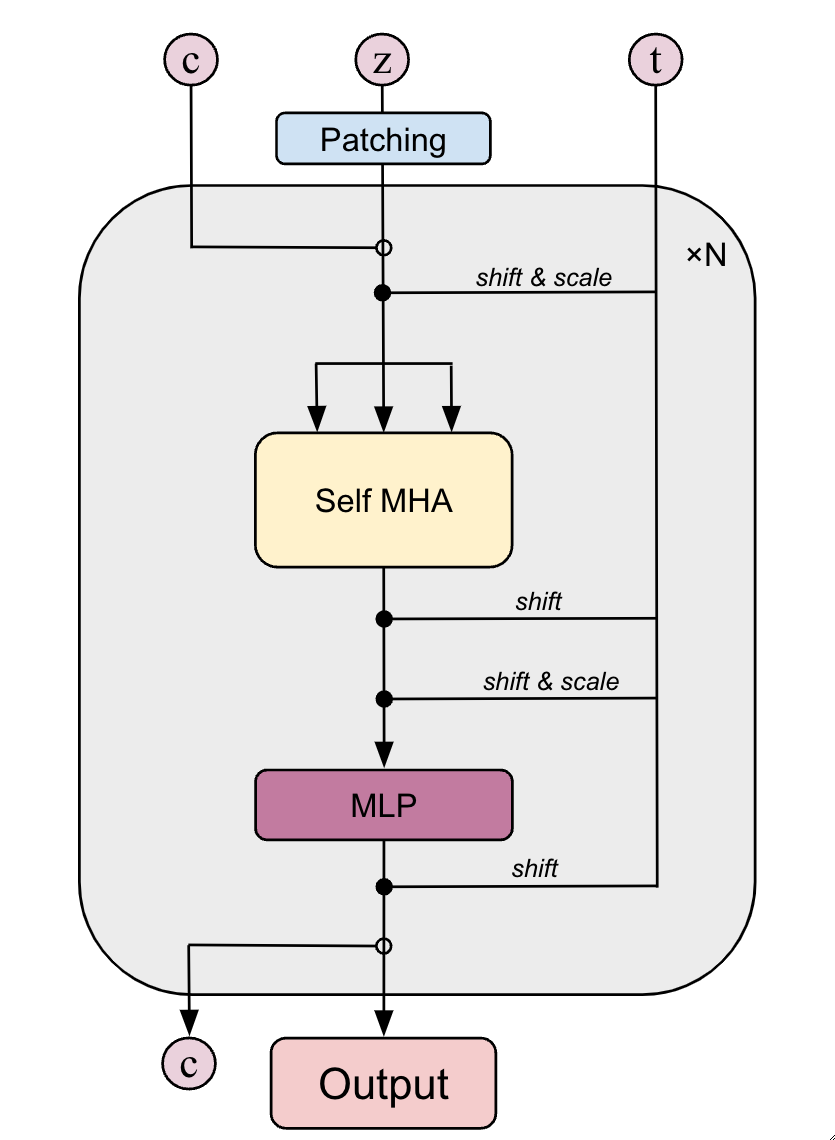} \\
        {\small (a) PixArt-\(\alpha\) } & 
        {\small (b) MMDiT } & 
        {\small (c) \model } \\
    \end{tabular}
    \caption{\textbf{Comparison of Diffusion Transformer Architectures.} Element-wise operations are denoted by $\bullet$, and sequence-wise operations by $\circ$. The details of inputs $\mathbf{c}$, $\mathbf{z}$, $t$ can be found in Figure~\ref{fig:latent_diffusion}. 
    PixArt-\(\alpha\) relies on sequential self- and cross-attention, whereas MMDiT uses a dual-stream approach with separate parameters for text and image tokens. Our proposed \model\ resembles a vanilla DiT that processes concatenated text and noises. \wliu{Should we put the concatenation of text (c) and noise (z) outside the N block? I assume we don't need to concatenate it in every block?} }
    \label{fig:architecture_comparison}
\end{figure*}

\paragraph{PixArt-\(\alpha\).}
PixArt-\(\alpha\) (Figure~\ref{fig:architecture_comparison}a) follows a two-step attention process: (1)~self-attention on patchified visual tokens, and (2)~cross-attention to fixed text embeddings \(\mathbf{c}\). These text embeddings remain the same across all layers, providing a global conditioning signal for generation.

\paragraph{MMDiT.}
MMDiT (Figure~\ref{fig:architecture_comparison}b) adopts a dual-stream approach in each transformer block: text and visual tokens have separate query, key, value, and output (QKVO) projections. After computing self-attention over the concatenated sequence, tokens from each modality are processed by separate MLPs, preserving modality-specific transformations. This design supports rich multimodal interactions but substantially increases the parameter count.

Table~\ref{tab:param_count} compares the main parameter components for PixArt-\(\alpha\) and MMDiT, highlighting that MMDiT consumes significantly more parameters due to:
\begin{itemize}
    \item Per-layer AdaLN: MMDiT uses independent AdaLN parameters in each layer, while PixArt-\(\alpha\) shares them.
    \item Dual-Stream Design: MMDiT duplicates QKVO and MLPs for text and image tokens, effectively doubling related parameter sets.
\end{itemize}

\begin{table}[t]
\centering
\caption{Parameter counts for PixArt-$\alpha$, MMDiT, \model, and \model-lite (full vs.\ QKVO).
$d$ represents the effective embedding size in transformer blocks (\ie, the model's hidden dimension), and $N$ denotes the number of layers. Our {\model} is a compact DiT with shared dual-stream  AdaLN, hence saves parameters significantly.} 
\scriptsize
\setlength{\tabcolsep}{2pt}
\resizebox{\columnwidth}{!}{%
\begin{tabular}{l|ccccc}
\toprule
\multirow{2}{*}{\textbf{Component}} 
  & \multirow{2}{*}{\textbf{PixArt-$\alpha$}}
  & \multirow{2}{*}{\textbf{MMDiT}}
  & \multirow{2}{*}{\textbf{\model}}
  & \multicolumn{2}{c}{\textbf{\model-Lite}} \\
  \cline{5-6}\noalign{\vspace{0.5ex}}
 &  &  & & 
 \textbf{(full)} & \textbf{(attention)} \\
\midrule
\textbf{AdaLN} 
  & $6d^2$
  & $12Nd^2$
  & $12d^2$
  & $12d^2$
  & $12d^2$ \\

\textbf{Self-MHA}
  & $4Nd^2$
  & $8Nd^2$
  & $4Nd^2$
  & $4d^2$
  & $4d^2$ \\

\textbf{Cross-MHA}
  & $4Nd^2$
  & --
  & --
  & --
  & -- \\

\textbf{MLP}
  & $8Nd^2$
  & $16Nd^2$
  & $8Nd^2$
  & $8d^2$
  & $8Nd^2$ \\

\textbf{Total}
  & $(6 + 16N)d^2$
  & $36Nd^2$
  & $(12 + 12N)d^2$
  & $24d^2$
  & $(16 + 8N)d^2$ \\
\bottomrule
\end{tabular}%
}

\label{tab:param_count}
\end{table}

\subsection{\model: A Compact Text-to-Image DiT}
In this section, we introduce \textbf{\model}, a compact DiT architecture tailored for text-conditioned generation. Instead of having the dual-stream approach of MMDiT, \model\  employs a unified set of QKVO projections and MLPs for both text and image tokens, offering a streamlined yet effective design (see Figure~\ref{fig:architecture_comparison}c).

%\model\ also retains PixArt-\(\alpha\)'s shared AdaLN parameters, 

For adaptive layer normalization (AdaLN), \model\ combines the dual-stream AdaLN from MMDiT with the parameter-sharing strategy of PixArt-\(\alpha\).  By sharing AdaLN parameters across all layers, \model\ effectively limits parameter growth as model depth increases, achieving a balanced trade-off between efficiency and multimodal capacity. Notably, this approach maintains a constant parameter overhead for AdaLN, irrespective of model depth.

As shown in Table \ref{tab:param_count}, \model's design reduces the parameter count by approximately \(24Nd^2\) compared to MMDiT, while preserving the same computational complexity (FLOPs). Overall, \model\ consumes about 66\% fewer parameters than MMDiT and 25\% fewer than PixArt-\(\alpha\) for large \(N\). 

However, two open questions remain:

\begin{itemize}
    \item \textit{Does sharing AdaLN across all layers adversely affect image quality?}
    \item \textit{How does merging text and image streams influence text alignment and fidelity?}
\end{itemize}

We investigate these aspects in Section~\ref{sec:exp}. Further, we also explore whether parameter savings can be pushed even more aggressively, as described next.

%-------------------------------------------------------------------------

\subsubsection{\model-Lite}
\label{sec:monodit-lite}

While {\model} strikes a balance between multimodal capacity and parameter efficiency, some applications (\eg, large-scale inference or deployment on resource-constrained hardware) demand even greater reductions in parameter count. Inspired by the shared AdaLN and ALBERT~\cite{lanalbert}, which shares parameters across layers to reduce model size, we propose DiT-Lite to further reduce parameters by sharing Transformer block parameters across layers, either entirely or partially (see Table~\ref{tab:param_count}).

\paragraph{Full Block-Sharing Variant.}
\label{sec:monodit-lite-full}
In this variant, the \emph{entire} Transformer block (QKVO projections and MLP) is shared across all $N$ layers. This reduces the overall parameter count to $\sim24d^2$, making the model extremely compact. However, it also limits representational diversity, since every layer applies the exact same transformation, which can degrade performance on complex prompts.

\paragraph{Attention-Sharing Variant.}
\label{sec:monodit-lite-qkvo}
A more moderate approach involves \emph{sharing only} the QKVO projections while maintaining a distinct MLP for each layer. As shown in Table~\ref{tab:param_count}, this partial-sharing strategy reduces the parameter count by approximately 33\% compared to the non-shared version, while generally preserving higher fidelity and better text alignment than the fully shared configuration. Having dedicated MLPs for each layer allows the model to capture depth-specific nuances in the text-to-image mapping.

% \paragraph{Further Considerations.}

% These MonoDiT-lite approaches showcase a flexible spectrum of parameter-sharing strategies for text-to-image generation, enabling more compact models without severely compromising quality. In Section~\ref{sec:exp}, we offer a comprehensive evaluation of PixArt-\(\alpha\), MMDiT, MonoDiT, and both MonoDiT-lite variants in terms of image fidelity, text alignment, and parameter usage.

% Finally, we also investigate how different text encoders (e.g., CLIP vs.\ T5) and VAEs influence generation. Section~\ref{sec:exp} discusses the interplay among backbone architecture, text encoders, and VAEs, presenting a holistic view of designing high-fidelity, parameter-efficient text-to-image diffusion models.

\section{Experiments}
\label{sec:exp}

In this section, we evaluate the impact of architectural variations and design choices using a standardized training and evaluation protocol. Specifically, by fixing other components (\eg, VAE, text encoder),  we isolate the effects of attention mechanisms and parameter utilization, allowing for a clear analysis of each architectural change.

\subsection{Experimental Setup}
We outline the dataset, training setup, model scaling strategy, and evaluation metrics used in our experiments.

% \begin{table*}[t]
% \centering
% \caption{Performance Comparison Between MMDiT Variants with and without Shared AdaLN.}
% \renewcommand{\arraystretch}{1.2} % Slightly increase row height
% \setlength{\tabcolsep}{3pt} % Adjust column spacing
% \small
% \begin{tabular}{l|c|ccccccc}
% \toprule[1.2pt]
% \textbf{Model} & \textbf{Params} & \textbf{Val Loss} $\downarrow$ & \textbf{FID} $\downarrow$ & \textbf{CLIP} $\uparrow$ & \textbf{Pick} $\uparrow$ & \textbf{Gen Eval} $\uparrow$ & \textbf{Aesth.} $\uparrow$ & \textbf{T2I Comp.} $\uparrow$ \\
% \midrule
% MMDiT/B$_{\text{w/o shared AdaLN}}$ & 902M & 0.4219 & 14.72 & 32.91 & 20.28 & 69.83 & 5.62 & 0.511 \\
% MMDiT/B$_{\text{w/ shared AdaLN}}$   & 631M & 0.4220 & 15.04 & 32.86 & 20.28 & 69.81 & 5.59 & 0.510 \\
% \bottomrule[1.2pt]
% \end{tabular}
% \label{tab:mmdits_comparison}
% \end{table*}

\begin{table}[t]
\centering
\caption{Performance comparison between MMDiT variants with and without sharing AdaLN. Both models are based on the MMDiT/B sized configuration.}
\renewcommand{\arraystretch}{1.2} % Slightly increase row height
\setlength{\tabcolsep}{3pt} % Adjust column spacing
\resizebox{\columnwidth}{!}{%
\begin{tabular}{l|c|ccccccc}
\toprule[1.2pt]
\textbf{Model} & \textbf{Params} & \textbf{Val.} $\downarrow$ & \textbf{FID} $\downarrow$ & \textbf{CLIP} $\uparrow$ & \textbf{Pick} $\uparrow$ & \textbf{GenE.} $\uparrow$ & \textbf{Aesth.} $\uparrow$ & \textbf{T2I.} $\uparrow$ \\
\midrule
\makecell[l]{Per-layer AdaLN} & 902M & 0.422 & 14.7 & 32.9 & 20.28 & 69.8 & 5.62 & 51.1 \\
\makecell[l]{Shared AdaLN}   & 631M & 0.422 & 15.0 & 32.9 & 20.28 & 69.8 & 5.59 & 51.0 \\
\bottomrule[1.2pt]
\end{tabular}%
}
\label{tab:mmdits_comparison}
\end{table}

%\begin{itemize}

%    \item 
\noindent\textbf{Data}:  
    We conduct all ablations on in-house data containing ~1.5 billion text-image pairs. Following DALL·E~3~\cite{betker2023improving}, we enrich the dataset using synthetic captions generated by a pretrained captioning model. To balance real and synthetic data, we adopt a 1:9 ratio between original and synthetic captions. Unlike prior work that often relies on smaller subsets, all ablation studies are performed on the full dataset, ensuring a more reliable assessment of real-world performance.

%    \item 
\noindent\textbf{Training and Inference}:  
    To ensure fair and consistent comparisons across experiments, we standardize key components: all models use a shared in-house variational autoencoder (VAE) and an in-house CLIP-H model as the default text encoder unless stated otherwise. Implementation details for these components are provided in the Appendix~\ref{sec:appendix_impl_details}. 
    
    All models are trained using a flow-matching objective, optimized with AdaFactor at a fixed learning rate of 1e-4, and a global batch size of 4096 for 1 million steps. For inference, we employ a Heun SDE solver~\cite{kidger2021on} with 50 sampling steps and a classifier-free guidance scale of 7.5. 
    
    For model scaling experiments, we evaluate models at five specifications: S, B, L, XL and XXL, which correspond to 12, 18, 24, 30, and 38 transformer layers, respectively. The hidden dimension $d$ scales proportionally with depth as $d = 64 \times n_\text{layer}$, following the scaling strategy of SD3~\cite{esser2024scaling}. Unless otherwise specified, all ablation studies use the B-size model. 
    % Add back when pushing public
    All experiments are done using the axlearn framework.\footnote{\url{https://github.com/apple/axlearn}}

%    \item 
\noindent\textbf{Evaluation}:  
   We evaluate model performance using a combination of validation loss and a diverse set of established benchmarks. Following SD3~\cite{esser2024scaling} and MovieGen~\cite{polyak2025moviegencastmedia}, we report validation loss on both our in-house dataset. While validation loss provides a general measure of model fit, it may not accurately capture text alignment performance, particularly in complex generative tasks. As discussed further in Section~\ref{sec: text encoder ablation}, this limitation underscores the importance of incorporating additional metrics that directly evaluate alignment and compositionality. Therefore, we include Fréchet Inception Distance~\cite{heusel2018ganstrainedtimescaleupdate} on COCO30k~\cite{lin2015microsoftcococommonobjects}, CLIPScore~\cite{radford2021learningtransferablevisualmodels, hessel2022clipscorereferencefreeevaluationmetric}, and PickScore~\cite{kirstain2023pickapicopendatasetuser} on MJHQ30k~\cite{li2024playgroundv25insightsenhancing}, along with GenEval~\cite{ghosh2023genevalobjectfocusedframeworkevaluating}, T2I CompBench~\cite{huang2023t2icompbenchcomprehensivebenchmarkopenworld} and LAION-Aesthetics Predictor V2 (Aesthetics)~\cite{schuhmann2022laion5bopenlargescaledataset}. For multi-category benchmarks such as GenEval and T2I CompBench, we report the overall average in the main paper with detailed per-category results deferred to the Appendix.

%\end{itemize}

\subsection{Adaptive Layer Normalization Sharing}
As {\model} can be viewed as a simplification of MMDiT by unifying text and image streams, we start with investigating the impact of sharing AdaLN parameters across layers on the quality of generated images by comparing two MMDiT variants: one with shared AdaLN parameters and one without. The results are presented in Table~\ref{tab:mmdits_comparison}.

\begin{figure}[t]
    \centering
    \includegraphics[width=1\linewidth]{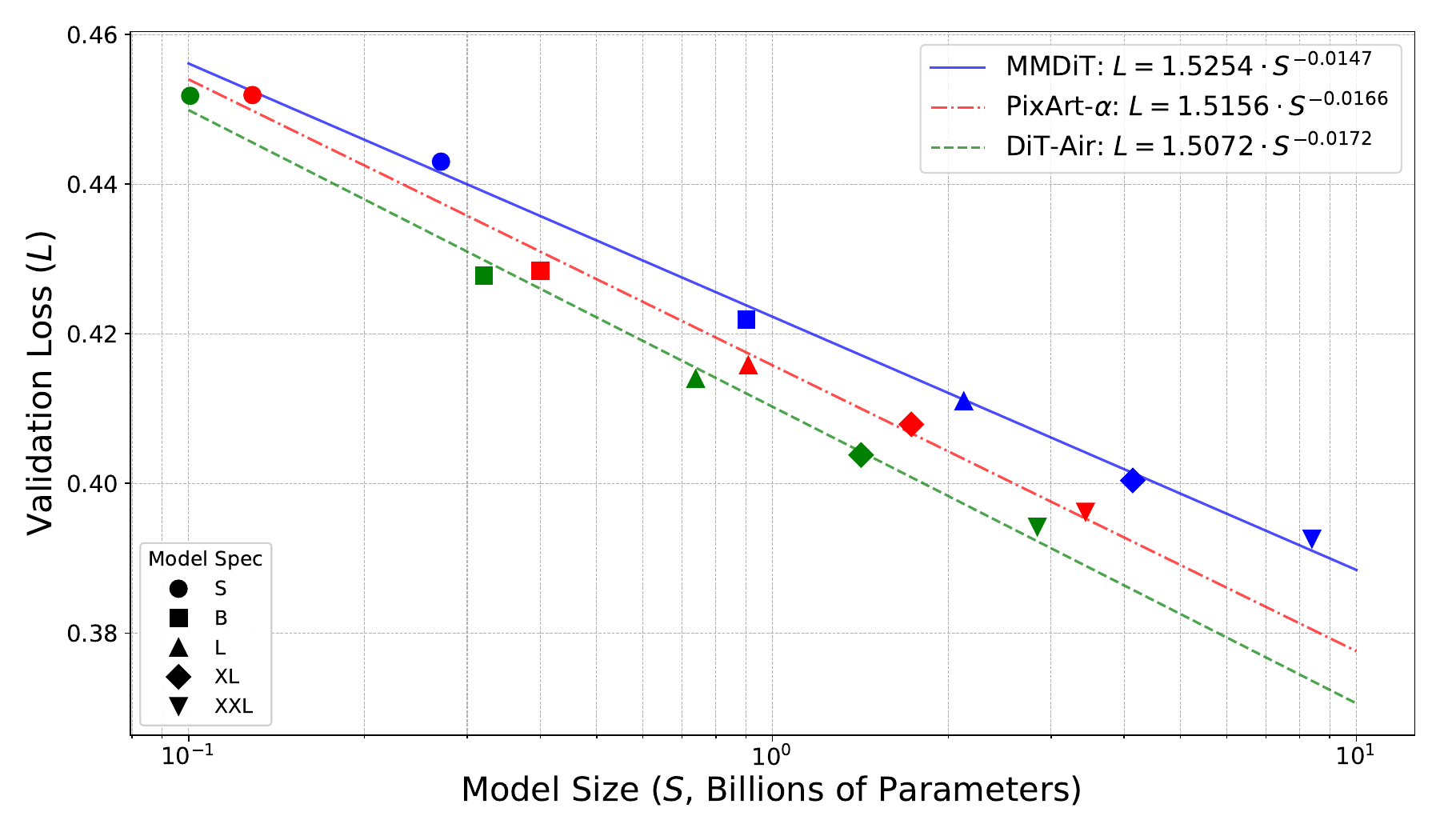}
    \caption{\textbf{Validation Loss vs.\ Model Size for PixArt-$\alpha$, MMDiT, and \model.} The plot illustrates the scaling behavior of three architectures across model sizes ranging from S to XXL, where the model size refers only to the diffusion transformer component (excluding the text encoder and VAE). The x-axis is in logarithmic scale, and the fitted lines depict the scaling trend using the formula \(L = a \cdot S^b\).  Among the three, \model\ achieves the best parameter efficiency.}
    \label{fig:validation_loss}
\end{figure}

\begin{figure*}[t]
    \centering
    \includegraphics[width=1.0\textwidth]{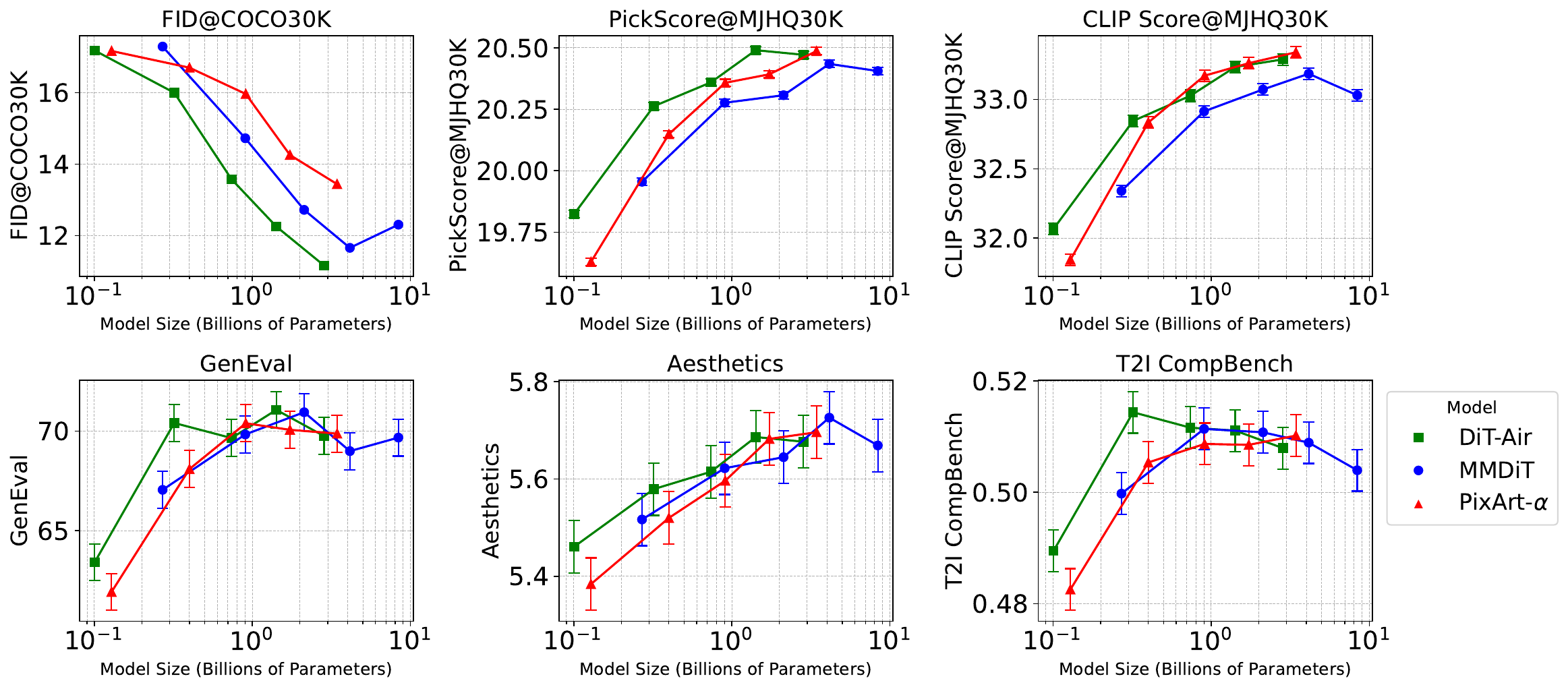}
    \caption{\textbf{Benchmark Performance Across Model Scales.} 
    The plots compare PixArt-$\alpha$, MMDiT, and \model\ across six evaluation metrics. \model\ demonstrates strong parameter efficiency, achieving competitive performance with fewer parameters. The x-axis is in logarithmic scale, and error bounds are indicated where applicable.}
    \label{fig:model_benchmark_comparison}
\end{figure*}

As shown, the variant with shared AdaLN parameters is more parameter-efficient, reducing the model size from 902M to 631M. The performance differences between the two variants are minimal, with only slight variations in validation loss and other evaluation metrics, all of which fall within the error bounds. Thus, we conclude that sharing AdaLN parameters improves parameter efficiency without significantly affecting image quality.

\subsection{Scaling and Efficiency}
\label{sec:scaling}
In this section, we evaluate the performance and efficiency of three architectures: PixArt-$\alpha$, MMDiT, and \model, for various model scales from S to XXL (150M to 8B parameters). Our analysis focuses on validation loss and benchmark performance to highlight the relationship between model size, parameter efficiency, and architecture.

\paragraph{Validation Loss and Scaling Behavior.}

Figure~\ref{fig:validation_loss} illustrates the validation loss versus model size for three architectures. \model\ demonstrates the best parameter efficiency, largely due to its shared AdaLN parameters across layers and its single-stream design for QKVO and MLPs. Among the three models, \model\ exhibits the steepest scaling curve, reflecting the most efficient reduction in validation loss as model size increases. Notably, at the S scale, \model’s validation loss is considerably higher than that of MMDiT; however, as the models scale up to XXL, the loss gap vanishes. PixArt-$\alpha$ follows a scaling trend similar to \model—with a slope substantially steeper than that of MMDiT—but for a fixed parameter budget, \model\ consistently achieves a lower validation loss, underscoring its balanced approach to parameter efficiency and performance.

\paragraph{Benchmark Performance.}

Figure~\ref{fig:model_benchmark_comparison} compares benchmark performance across all three architectures and model scales. Overall, \textbf{\model} demonstrates consistently strong performance with a significantly higher degree of parameter efficiency. Notably, \model\ achieves some of the lowest FID scores while performing on par or slightly better than MMDiT and PixArt-$\alpha$ in PickScore and GenEval. Although MMDiT exhibits higher aesthetics scores at larger scales and PixArt-$\alpha$ achieves strong CLIPScores, these gains often remain within the error bounds or necessitate considerably more parameters. This highlights \model’s strength in maintaining competitive performance without the overhead of excessive scaling.

No single metric fully captures all aspects of text-to-image alignment and quality, underscoring the importance of evaluating models across a diverse set of benchmarks. Overall, the results establish \model\ as a compelling choice for large-scale text-to-image generation, offering a well-balanced mix of efficiency and performance.

\subsection{\model-Lite Ablation}
\label{sec:monodit-lite-ablation}

Having established in Section~\ref{sec:scaling} that \model~(DiT with shared AdaLN) compares favorably to PixArt-$\alpha$ and MMDiT, we now evaluate \model-Lite, the more aggressive parameter-sharing extension introduced in Section~\ref{sec:monodit-lite}. We consider two configurations: \emph{Full Block-Sharing} and \emph{Attention-Sharing}. Full Block-Sharing reuses entire block across all layers, whereas Attention-Sharing keeps distinct MLPs per layer but shares QKVO. Table~\ref{tab:monodit_lite_comparison} compares both \model-lite variants against the baseline.

% \begin{table*}[t]
% \centering
% \caption{Comparison of MonoDiT (baseline) with MonoDiT-lite (full) and MonoDiT-lite (attention). The \emph{Full} configuration provides the largest parameter reduction but suffers a noticeable performance drop, while \emph{Attention} strikes a more favorable trade-off between compactness and text alignment.}
% \renewcommand{\arraystretch}{1.2} % Slightly increase row height
% \setlength{\tabcolsep}{3pt} % Adjust column spacing
% \small
% \begin{tabular}{l|c|ccccccc}
% \toprule[1.2pt]
% \textbf{Model} & \textbf{Params} & \textbf{Val Loss} $\downarrow$ & \textbf{FID} $\downarrow$ & \textbf{CLIP} $\uparrow$ & \textbf{Pick} $\uparrow$ & \textbf{Gen Eval} $\uparrow$ & \textbf{Aesth.} $\uparrow$ & \textbf{T2I Comp.} $\uparrow$ \\
% \midrule
% MonoDiT/B                     & 321M & 0.4278 & 16.00 & 32.84 & 20.26 & 70.41 & 5.58 & 0.514 \\
% MonoDiT/B-lite (full)         & 49M  & - & - & - & - & - & - & - \\
% MonoDiT/B-lite (attention)    & 230M & - & - & - & - & - & - & - \\
% \bottomrule[1.2pt]
% \end{tabular}
% \label{tab:monodit_lite_comparison}
% \end{table*}

\begin{table}[t]
\centering
\caption{Comparison of \model\ (baseline) with \model-lite (full) and \model-lite (attention). The \emph{Full} configuration provides the largest parameter reduction but suffers a noticeable performance drop, while \emph{Attention} strikes a more favorable trade-off between compactness and text alignment.}
\renewcommand{\arraystretch}{1.2} % Slightly increase row height
\setlength{\tabcolsep}{3pt} % Adjust column spacing
\resizebox{\columnwidth}{!}{%
\begin{tabular}{l|r|ccccccc}
\toprule[1.2pt]
\textbf{Model} & \textbf{Params} & \textbf{Val.} $\downarrow$ & \textbf{FID} $\downarrow$ & \textbf{CLIP} $\uparrow$ & \textbf{Pick} $\uparrow$ & \textbf{GenE.} $\uparrow$ & \textbf{Aesth.} $\uparrow$ & \textbf{T2I.} $\uparrow$ \\
\midrule
\makecell[l]{\model/B} & 321M & 0.428 & 16.0 & 32.8 & 20.2 & 70.4 & 5.58 & 51.4 \\
\makecell[l]{\model/B-lite \\ \footnotesize(full)}   & 49M  & 0.461 & 14.8 & 31.4 & 19.5 & 58.4 & 5.36 & 47.6 \\
\makecell[l]{\model/B-lite \\ \footnotesize(attention)}   & 230M  & 0.431 & 17.4 & 32.5 & 20.1 & 66.9 & 5.50 & 49.6 \\
\bottomrule[1.2pt]
\end{tabular}%
}
\label{tab:monodit_lite_comparison}
\end{table}

These results confirm that sharing the entire block minimizes parameters aggressively, but at the cost of lower text alignment and aesthetics. In contrast, \emph{attention}-only sharing still yields substantial parameter savings while incurring only modest performance drops. Overall, \model-lite (attention) emerges as a favorable option when computational and memory constraints are critical, yet text alignment and image quality must remain high.

\section{Text Encoders and VAEs}
In this section we further investigate the impact of other critical components in the text-conditioned image generation task: text encoders and VAEs.

\subsection{Text Encoder Ablation}
\label{sec: text encoder ablation}

We investigate how different text encoders impact the text alignment and overall performance of the {\model} architecture. Our study evaluates three primary encoder types: CLIP, T5, and Large Language Models (LLMs). Both the CLIP and LLM models are internal implementations, while the T5 encoder utilizes the open-sourced T5-XXL model. The analysis includes ablations of causal vs.\ bidirectional CLIP, layer selection strategies for both CLIP and LLMs, and a final comparison of all three encoders. Detailed ablation results are provided in  Appendix~\ref{sec:appendix_text_encoder_ablation}.

\paragraph{Summary of Key Findings.}

Our experiments demonstrate that the bidirectional CLIP model consistently outperforms its causal counterpart, showing improved text alignment and image quality across benchmarks. This improvement is attributed to the better synergy between the bidirectional attention in the text encoder and the diffuison transformer. Layer selection experiments, detailed in  Appendix~\ref{sec:appendix_clip_layer}, indicate that while very shallow layers in the CLIP model underperform, deeper layers yield comparable results.\footnote{In this paper, "shallow" layers always refer to those near the input of the text encoder, while "deep" layers are closer to the output, with depth consistently measured from input to output.}

For LLM-based encoders, we find that text-only LLMs outperform their multimodal counterparts, particularly in the GenEval metric. Layer selection studies in LLMs reveal that using a middle layer combined with a layer at approximately 3/4 of the model's depth offers the best performance. This behavior likely stems from the LLM's pretraining objective, which is typically next-token prediction. Such an objective emphasizes fine-grained, token-level information at deeper layers, potentially at the expense of broader semantic understanding, which is more beneficial for text-to-image alignment.

\paragraph{Comparison of CLIP, LLM, and T5.}

% \begin{table*}[t]
% \centering
% \caption{Summary of Text Encoder Performance Across Key Metrics. Detailed ablation results are available in the Appendix.}
% \renewcommand{\arraystretch}{1.2} % Slightly increase row height
% \setlength{\tabcolsep}{3pt} % Adjust column spacing
% \small
% \begin{tabular}{l|cccccccc}
% \toprule[1.2pt]
% \textbf{Text Encoder} & \textbf{Val Loss} $\downarrow$ & \textbf{FID} $\downarrow$ & \textbf{CLIP} $\uparrow$ & \textbf{Pick} $\uparrow$ & \textbf{Gen Eval} $\uparrow$ & \textbf{Aesth.} $\uparrow$ & \textbf{T2I Comp.} $\uparrow$ \\
% \midrule
% CLIP (Bidirectional) & 0.4278 & 16.00 & \textbf{32.84} & \textbf{20.26} & 0.704 & \textbf{5.579} & \textbf{0.514} \\
% LLM (Text-only) & 0.4269 & \textbf{15.99} & 31.99 & 20.09 & \textbf{0.726} & 5.567 & 0.486 \\
% T5-XXL & \textbf{0.4244} & 17.45 & 31.77 & 19.96 & 0.653 & 5.460 & 0.480 \\
% \bottomrule[1.2pt]
% \end{tabular}
% \label{tab:text_encoder_summary}
% \end{table*}

\begin{table}[t]
\centering
\caption{Comparison of CLIP, LLM, and T5 text embeddings. CLIP achieves superior results across most metrics, while LLM excels in GenEval. In contrast, T5 consistently underperforms.}
\renewcommand{\arraystretch}{1.2} % Slightly increase row height
\setlength{\tabcolsep}{3pt} % Adjust column spacing
\resizebox{\columnwidth}{!}{%
\begin{tabular}{l|c|ccccccc}
\toprule[1.2pt]
\textbf{Model} & \textbf{Val.} $\downarrow$ & \textbf{FID} $\downarrow$ & \textbf{CLIP} $\uparrow$ & \textbf{Pick} $\uparrow$ & \textbf{GenE.} $\uparrow$ & \textbf{Aesth.} $\uparrow$ & \textbf{T2I.} $\uparrow$ \\
\midrule
CLIP (Bidirectional) & 0.428 & 16.0 & \textbf{32.8} & \textbf{20.3} & 0.704 & \textbf{5.58} & \textbf{51.4} \\
LLM (Text-only) & 0.427 & \textbf{16.0} & 32.0 & 20.1 & \textbf{0.726} & 5.57 & 48.6 \\
T5-XXL & \textbf{0.424} & 17.5 & 31.8 & 20.0 & 0.653 & 5.46 & 48.0 \\
\bottomrule[1pt]
\end{tabular}%
}
\label{tab:text_encoder_summary}
\end{table}

The final comparison of text encoders (Table~\ref{tab:text_encoder_summary}) shows that bidirectional CLIP achieves the best performance across most benchmarks, with text-based LLMs also performing strongly, especially in GenEval. The T5-XXL model, while achieving the lowest validation loss, generally lags behind in benchmark performance. This discrepancy highlights an important observation: validation loss alone is not always indicative of text alignment performance, particularly when comparing different text encoder architectures. Models with lower validation loss, such as T5, may not necessarily offer the best real-world performance on established benchmarks.

\paragraph{Takeaway for Final Model Design.}

Based on these findings, our final model adopts a hybrid strategy, combining bidirectional CLIP with a text-based LLM to leverage both efficient text alignment and deeper semantic understanding. This approach ensures a balanced trade-off between parameter efficiency and robust results across diverse metrics.

% put here for page alignment
\begin{table*}[ht]
\centering
\caption{Comparison with state-of-the-art (SoTA) models. ``Total (B)'' includes parameters from the text encoder, VAE, and diffusion model, while ``Trainable (B)'' denotes only those parameters updated during training. The value marked with $\dagger$ is estimated from Figure~8 in~\cite{esser2024scaling} as the exact number was unavailable.}
\renewcommand{\arraystretch}{1.2}
\setlength{\tabcolsep}{8pt} % Increase column spacing for readability
\begin{tabular}{l|rr|rccccl}
\toprule[1.2pt]
\multirow{2}{*}{\textbf{Model}} 
& \multicolumn{2}{c|}{\textbf{Size (B)}} 
& \multicolumn{6}{c}{\textbf{Metrics}} \\
\cmidrule(lr){2-3} \cmidrule(lr){4-9}
& \textbf{Total} & \textbf{Trainable} 
& \textbf{FID} $\downarrow$
& \textbf{CLIP} $\uparrow$ 
& \textbf{Pick} $\uparrow$ 
& \textbf{GenE.} $\uparrow$ 
& \textbf{Aesth.} $\uparrow$ 
& \textbf{T2I.} $\uparrow$ \\
\midrule
SDXL Base~\cite{podell2023sdxl}    & 3.5 & 2.6 & 268.0 & 22.1 & 17.0 & 55.0 & 4.32 & 40.6 \\
PixArt-$\alpha$~\cite{chenpixart} & 5.4 & 0.6 & 120.7 & 27.3 & 17.0 & 55.7 & 5.76 & 44.7 \\
SD3 Medium~\cite{esser2024scaling}& 7.7 & 2.0 & 26.0  & 32.0 & 20.7 & 62.0 & 5.99 & 52.4\\
SD3~\cite{esser2024scaling}       & 13.6 & 8.0 & -- & -- & -- & 74.5 & -- & 51.4$^\dagger$ \\
Flux-Dev~\cite{flux2024}          & 16.9 & 12.0 & 68.7 & 30.2 & 19.7 & 66.7 & 6.12 & 49.6 \\
Flux-Schnell~\cite{flux2024}      & 16.9 & 12.0 & 25.1 & 33.1 & 21.6 & 70.7 & 6.12 & 49.9 \\
JanusPro-7B~\cite{chen2025janus}  & 6.9 & 6.9  & \textbf{17.2} & 15.5 & 16.6 & 80.3 & 5.95 & 35.2 \\
\midrule
\model/L-Lite                     & 1.2 & 0.7 & 23.1 & 33.9 & 21.5 & 78.4 & 6.06 & 55.4 \\
\model/XXL                        & 6.0 & 2.8 & 32.2 & \textbf{34.7} & \textbf{22.1} & \textbf{82.9} & \textbf{6.29} & \textbf{59.5} \\
\bottomrule[1.2pt]
\end{tabular}
\label{tab:sota_comparison}
\end{table*}

\subsection{Progressive VAE Training}
\label{sec: improved vae}

The selection of a Variational Autoencoder (VAE) and its training strategy is critical for achieving high image fidelity in text-to-image generation. Although increasing the channel capacity of the VAE generally improves image reconstruction quality, it can inflate the KL divergence, hindering subsequent diffusion training.

To balance this trade-off, we introduce a progressive training pipeline. Our approach starts by training a low-channel VAE (\eg, with 4 channels) from scratch. In a second stage, an intermediate convolutional layer is replaced with a higher channel capacity (\eg, upgraded to 8 channels), followed by continued training. Using this two-stage process, we trained an 8-channel VAE that is employed in all our models. This method not only enhances downstream text-to-image generation but also maintains competitive reconstruction performance compared to the 4-channel or 8-channel VAEs trained from scratch. 

Our ablation studies reveal that training an 8-channel VAE from scratch achieves an rFID of 2.59, whereas our progressive approach—starting with a 4-channel VAE and later expanding to 8 channels—attains a comparable rFID of 2.61 while reducing the KL divergence from 9$\times10^5$ to 7$\times10^4$. This reduction in divergence is critical, as it leads to improved downstream text-to-image performance (GenEval of 70.4 and T2I CompBench of 51.4) compared to 69.4 and 50.7 when using the 8-channel model trained from scratch. More details and comparisons are provided in Appendix~\ref{sec:appendix vae}.

\section{Final Models}
\label{sec:final_model}

In this section, we introduce our two final models, \textbf{\model/XXL} and \textbf{\model/L-Lite (attention)}, developed through our progressive training pipeline. Both models undergo a multi-stage training process: initial training at 
 \(256^2\) resolution, further training at  \(512^2\), followed by supervised fine-tuning (SFT) on a curated subset following \citet{dai2023emu}, and finally refined with reward fine-tuning using an approach similar to DRaFT~\cite{kevin2024directlyfinetuningdiffusionmodels} using the HPSv2~\cite{xiaoshi2023humanpreferencescorev2} reward model.\footnote{Detailed training procedures are provided in Appendix~\ref{sec:appendix training_recipe} (Configurations), \ref{sec:appendix sft} (SFT), and \ref{sec:appendix rl} (Reward fine-tuning).}

\paragraph{\model/XXL.}
\model/XXL represents our high-capacity model, combining a bidirectional CLIP text encoder  for efficient text alignment and a text-based LLM  for rich semantic understanding, as detailed in Section~\ref{sec: text encoder ablation}.  We opt for the XXL configuration to further push the boundaries of image quality.

\paragraph{\model/L-Lite (attention).}
For a more parameter-efficient solution, we introduce \model/L-Lite (attention), which relies solely on the bidirectional CLIP text encoder and an L-sized architecture. This design has a total of \textbf{1.15B} parameters—including the text encoder, VAE, and diffusion transformer—while maintaining competitive performance.

Image generation results for both models are provided in Appendix~\ref{sec:appendix_samples}.

\paragraph{Comparison with State-of-the-Art.}

As shown in Table~\ref{tab:sota_comparison}, our models compare favorably with current state-of-the-art text-to-image systems. In particular, \model/XXL achieves an exceptional GenEval overall score of \textbf{82.9} and a T2I CompBench average score of \textbf{59.5}, outperforming many larger competitors while maintaining a relatively compact size of 5.95B parameters. Meanwhile, \model/L-Lite offers a compelling balance between efficiency and quality, with strong performance across metrics such as CLIPScore, PickScore, and T2I CompBench scores. Although FID is included for consistency, we note that its reliability diminishes post fine-tuning due to distribution shifts.

\paragraph{Multi-stage Training.}

\begin{table}[t]
    \centering
    \caption{Comparison of different training strategies. Resolutions (256\(^2\) vs.\ 512\(^2\)) refer to the image size used during pretraining.}
    \renewcommand{\arraystretch}{1.2}
    \setlength{\tabcolsep}{3pt}
    \resizebox{\columnwidth}{!}{%
    \begin{tabular}{l|rrrrrr}
        \toprule[1.2pt]
        \textbf{Training Stage} & \textbf{FID} $\downarrow$ & \textbf{CLIP} $\uparrow$ & \textbf{Pick} $\uparrow$ & \textbf{GenE.} $\uparrow$ & \textbf{Aesth.} $\uparrow$ & \textbf{T2I.} $\uparrow$ \\
        \midrule
        Pretrain 256\(^2\)       & 12.0 & 33.4 & 20.5 & 71.1 & 5.57 & 50.7 \\
        Pretrain 512\(^2\)       & 13.0 & 33.5 & 20.7 & 74.2 & 5.62 & 51.7 \\
        Supervised fine-tuning                     & 22.5 & 34.2 & 21.5 & 79.0 & 5.89 & 55.3 \\
        Reward fine-tuning      & 32.2 & 34.7 & 22.1 & 82.9 & 6.21 & 59.5 \\
        \bottomrule[1.2pt]
    \end{tabular}%
    }
    \label{tab:multi_stage_training}
\end{table}

Table~\ref{tab:multi_stage_training} details the metrics of the multi-stage training pipeline for \model/XXL. Increasing the pretraining resolution from \(256^2\) to \(512^2\) slightly increases FID, indicating that higher-resolution data alone does not fundamentally alter the model’s capacity for coherent image generation under similar training conditions. At the same time, other metrics improve modestly, suggesting that a larger input resolution helps capture finer semantic and aesthetic details before any fine-tuning. Subsequent supervised and reward-based fine-tuning produce a more noticeable shift in FID—likely driven by distribution changes introduced by specialized or narrower data—yet these stages yield a marked improvement in text–image alignment and overall quality. 

\section{Conclusion}

In this work, we systematically study the performance of Diffusion Transformers (DiTs) for text-to-image generation, focusing on architectural choices, text-conditioning strategies, and training protocols. We find that the standard DiT architecture, when enhanced with shared AdaLN parameters and configured to directly process concatenated text and noise inputs, achieves superior parameter efficiency compared to alternative approaches, particularly at scale.

We introduce \textbf{\model} and \textbf{\model-Lite}, two models that enhance the parameter efficiency of the standard DiT backbone while carefully balancing model size and performance for text-to-image generation. Through comprehensive ablation studies of text encoders and variational autoencoders (VAEs), we identified design choices that significantly improve text-conditioned image generation quality. By applying a multi-stage training process—including supervised and reward fine-tuning—our final models set new state-of-the-art performance across key text-to-image generation benchmarks, outperforming existing models.

Our findings offer valuable insights into the development of more efficient and expressive text-to-image models, underscoring the potential for further optimizing diffusion architectures and training practices.

\section*{Acknowledgements}

We extend our sincere thanks to Jiasen Lu, Zhe Gan, Liangchen Song, Saeed Khorram and Pengsheng Guo whose constructive discussions and timely feedback propelled our research forward. We appreciate Vasileios Saveris, Jeff Lai, Aman Agarwal and Shubham Gupta for their early efforts in data preparation and processing, which lays a solid foundation for this work. We also gratefully acknowledge the dedicated infrastructure support provided by the Apple Foundation Model team, enabling seamless experimentation and data analysis. Finally, we are deeply indebted to the leadership from Ruoming Pang and Yang Zhao for their guidance, vision, and unwavering support throughout this project.
{
    \small
    \bibliographystyle{ieeenat_fullname}
    \bibliography{main}
}
\clearpage
\setcounter{page}{1}
\maketitlesupplementary

\appendix

\section{Implementation Details}
\label{sec:appendix_impl_details}

Our model builds upon a diffusion transformer framework with design choices that enhance training stability and performance.

\subsection{Model}

\subsubsection{Diffusion Transformer Variants}
We define five variants of the diffusion transformer: S, B, L, XL, and XXL. These correspond to models with 12, 18, 24, 30, and 38 transformer layers, respectively. The hidden dimension scales proportionally with the number of layers as
\(
d = 64 \times n_{\text{layer}},
\)
and the number of attention heads is set to equal the transformer depth. 

\subsubsection{Stability Enhancements}
To improve training stability, we integrate several techniques. First, we apply QK-normalization~\cite{dehghani2023scaling} following the SD3 methodology to stabilize query-key interactions within the attention mechanism. We also employ sandwich normalization~\cite{gong2022sandwich-norm,ding2021cogview} in both the attention blocks and MLP modules, a method proven effective for large-scale model training. Additionally, rather than using static positional embeddings for visual tokens, we incorporate a 2D rotary positional embedding~\cite{su2024roformer} within the attention mechanism to dynamically capture spatial relationships.

\subsubsection{Conditional Inputs via AdaLN}
In our implementation, we adopt the MMDiT approach by incorporating a pooled text embedding through Adaptive Layer Normalization (AdaLN) alongside the time embedding. We believe that this pooled embedding provides high-level semantic information with only a minimal increase in computational cost.

\subsubsection{Textual and Language Components}
For the textual component, we utilize an internal CLIP/H text encoder consisting of 24 transformer layers, a hidden dimension of 1024, and 16 attention heads, totaling approximately 335 million parameters. The pooled embedding is generated via last token pooling for causal CLIP or average pooling for bidirectional CLIP. Additionally, our internal language model (LLM) consists of 56 transformer layers, a hidden dimension of 6,656, and 16 attention heads, with around 2.8 billion parameters, and employs last token pooling to produce the pooled embedding for AdaLN.

\subsection{Training and Inference Details}
\label{sec:training_inference}

\subsubsection{Training Objective}
The model is trained using a \emph{flow-matching} objective:
\[
    \min_\theta \mathbb{E}_{(\mathbf{z}_0,\mathbf{c}),\, \epsilon \sim \mathcal{N}(0, I),\, t \sim \mathcal{P}(t)} \left[ \left\| f_\theta(\mathbf{z}_t, \mathbf{c}, t) - \mathbf{z}_0 + \epsilon \right\|_2^2 \right].
\]
In this formulation, the timestep distribution \(\mathcal{P}(t)\) follows a logit-Normal distribution as in SD3~\cite{esser2024scaling}, which emphasizes intermediate steps during the flow-matching process.

\subsubsection{Training Setup}
\label{sec:appendix training_recipe}
Training is conducted on TPU v5p hardware. We use the AdaFactor optimizer with a constant learning rate of \(1\times10^{-4}\), and momentum parameters \(b_1=0.9\) and \(b_2=0.999\).

For ablation studies, the model is trained for 1 million steps with a batch size of 4,096 at a resolution of \(256^2\). In contrast, the final models are trained in multiple stages:
\begin{enumerate}
    \item An initial stage of 500k steps at \(256^2\) resolution with a batch size of 4,096.
    \item A subsequent stage of 100k steps at \(512^2\) resolution with a batch size of 2,048.
    \item A supervised fine-tuning (SFT) stage for 2.5k steps with a batch size of 64, where the timestep distribution \(\mathcal{P}(t)\) is shifted so that the log-SNR aligns with that of the low-resolution training.
    \item A reward fine-tuning stage for an additional 4.8k steps with a batch size of 64.
\end{enumerate}
Further details regarding the SFT and reward finetuning stages are provided in Appendix~\ref{sec:appendix sft} and Appendix~\ref{sec:appendix rl}, respectively.

\subsubsection{Inference}
During inference, we employ a second-order Heun SDE solver~\cite{kidger2021on} with 50 sampling steps, combined with a classifier-free guidance scale of 7.5 to steer the sampling process.

\section{Evaluation}
\label{sec:appendix_eval}

We evaluate model performance using a combination of validation loss and several established benchmarks.

\subsection{Validation Loss}
Recent works such as SD3~\cite{esser2024scaling} and MovieGen~\cite{polyak2025moviegencastmedia} have proposed using validation loss as a performance estimate. We follow this trend. More specifically, in a flow-matching paradigm, the validation loss measures how well the model learns the velocity field induced by the transport equation. With a fixed VAE across all experiments, the source and target distributions remain constant (\ie, the standard normal distribution and the VAE-encoded latents, respectively). Consequently, the validation loss directly reflects the model's ability to predict the rectified flow trajectory, which often aligns with human perceptual preferences.

\subsection{Benchmark Metrics}
We report performance on the following metrics:
\begin{itemize}
    \item \textbf{Fréchet Inception Distance (FID):} Quantifies the similarity between the generated and real image distributions by comparing their Inception-v3 embeddings~\cite{heusel2018ganstrainedtimescaleupdate}.
    \item \textbf{CLIPScore:} Assesses the semantic alignment between images and text using CLIP embeddings. Higher scores denote better correspondence~\cite{radford2021learningtransferablevisualmodels, hessel2022clipscorereferencefreeevaluationmetric}.
    \item \textbf{PickScore:} Similar to CLIPScore, but based on a CLIP model trained on an open dataset of text-to-image prompts and real user preferences, thereby achieving compelling performance in predicting human preferences~\cite{kirstain2023pickapicopendatasetuser}.
    \item \textbf{GenEval:} Provides an overall evaluation of image generation performance. The original implementation, using 4 samples per prompt, tends to exhibit larger uncertainty bounds; to mitigate this, we increase the number of samples to 64~\cite{ghosh2023genevalobjectfocusedframeworkevaluating}.
    \item \textbf{T2I CompBench:} A comprehensive benchmark for assessing text-to-image synthesis quality~\cite{huang2023t2icompbenchcomprehensivebenchmarkopenworld}.
    \item \textbf{LAION-Aesthetics Predictor V2:} Predicts the aesthetic quality of images, with higher scores indicating superior visual appeal~\cite{schuhmann2022laion5bopenlargescaledataset}. In our evaluation, images are generated using ImageReward prompts~\cite{xu2023imagereward} and subsequently assessed with this aesthetic model.
\end{itemize}

% For state-of-the-art (SoTA) models, we use GenEval from~\cite{xie2024sana} and T2I CompBench from~\cite{huang2025t2i}; all other metrics were computed in-house.

\subsection{Abbreviation Key}
\label{sec:appendix_eval_abbreviation}
For clarity, we list the abbreviations used in tables throughout the paper:
\begin{itemize}
    \item \textbf{Val.} --- Validation Loss.
    \item \textbf{FID} --- Fréchet Inception Distance.
    \item \textbf{CLIP} --- CLIPScore.
    \item \textbf{Pick} --- PickScore.
    \item \textbf{GenE.} --- GenEval.
    \item \textbf{Aesth.} --- LAION-Aesthetics Predictor V2.
    \item \textbf{T2I.} --- T2I CompBench.
\end{itemize}

\section{Detailed Text Encoder Ablation Studies}
\label{sec:appendix_text_encoder_ablation}

This section provides a detailed analysis of the text encoder ablation experiments, including the impact of causal vs.\ bidirectional attention in CLIP, the effect of layer selection in both CLIP and LLMs, and a comparison between text LLM and multi-modal LLM. The results presented here supplement the summary findings discussed in Section~\ref{sec: text encoder ablation} of the main paper.

\subsection{CLIP}
\subsubsection{Causal vs.\ Bidirectional Attention in CLIP}
\label{sec:appendix_clip_layer}

\begin{table}[t]
\centering
\caption{Zero-Shot performance of causal vs.\ bidirectional CLIP models on ImageNet and COCO5k.}
\renewcommand{\arraystretch}{1.2}
\setlength{\tabcolsep}{3pt}
\resizebox{\columnwidth}{!}
{%
\begin{tabular}{l|cc|cccc}
\toprule[1.2pt]
\multirow{2}{*}{\textbf{ CLIP Model}} & \multicolumn{2}{c|}{\textbf{ImageNet}} & \multicolumn{4}{c}{\textbf{COCO5k}} \\ \cmidrule(lr){2-3} \cmidrule(lr){4-7}
 & \textbf{Acc@1} $\uparrow$ & \textbf{Acc@5} $\uparrow$ & \textbf{I2T@1} $\uparrow$ & \textbf{I2T@5} $\uparrow$ & \textbf{T2I@1} $\uparrow$ & \textbf{T2I@5} $\uparrow$ \\
\midrule
\makecell{Causal}    & 80.6 & 96.5 & 74.4 & 91.5 & 53.6 & 77.7 \\
\makecell{Bidirectional} & 80.6 & 96.5 & 74.6 & 91.5 & 53.8 & 78.3 \\
\bottomrule[1.2pt]
\end{tabular}%
}
\label{tab:clip_zero_shot}
\end{table}

\begin{table}[t]
\centering
\caption{Performance comparison of causal vs.\ bidirectional CLIP models as text embedding models for text-to-image generation.}
\renewcommand{\arraystretch}{1.2}
\setlength{\tabcolsep}{3pt}
\resizebox{\columnwidth}{!}{%
\begin{tabular}{c|ccccccc}
\toprule[1.2pt]
\textbf{CLIP Model} & \textbf{Val.} $\downarrow$ & \textbf{FID} $\downarrow$ & \textbf{CLIP} $\uparrow$ & \textbf{Pick} $\uparrow$ & \textbf{GenE.} $\uparrow$ & \textbf{Aesth.} $\uparrow$ & \textbf{T2I.} $\uparrow$ \\
\midrule
\makecell{Causal}     & 0.429 & 16.4 & 32.8 & 20.2 & 0.683 & 5.61 & 50.6 \\
\makecell{Bidirectional} & \textbf{0.428} & \textbf{16.0} & 32.8 & \textbf{20.3} & \textbf{0.704} & \textbf{5.58} & \textbf{51.4} \\
\bottomrule[1.2pt]
\end{tabular}%
}
\label{tab:causal_vs_bidirectional_clip}
\end{table}

To assess the impact of attention mechanisms in the text encoder, we compared causal and bidirectional variants of the CLIP/H model. In the causal configuration, the CLIP model employs causal attention with last-token pooling, whereas the bidirectional variant uses global average pooling during contrastive loss training. Both models exhibit comparable performance in standard zero-shot classification and retrieval tasks, as summarized in Table~\ref{tab:clip_zero_shot}. However, as shown in Table~\ref{tab:causal_vs_bidirectional_clip}, the bidirectional CLIP consistently outperforms its causal counterpart in terms of text alignment and image quality benchmarks. We hypothesize that the observed improvements in diffusion models arise specifically from enhanced attention alignment, rather than from intrinsic differences in the pretrained text encoder performance.

\subsubsection{Layer Selection in CLIP}

We investigated how the selection of different layers in the bidirectional CLIP text encoder (24 layers) affects performance. Embeddings from layers 6, 12, 18, 23, and 24 were tested, along with a concatenation of multiple layers (6, 12, 18, 24) followed by a linear projection. The 23rd layer was included as part of this study due to its common use as the penultimate layer in open-source text-to-image models. As shown in Table~\ref{tab:clip_layer_ablation}, our results indicate that all deeper layers (12, 18, 23, 24) exhibit comparable performance, while the shallow layer (6) underperforms. Concatenating embeddings from multiple layers did not yield significant improvements, suggesting that a single mid-to-deep layer is sufficient for robust text alignment.

\begin{table}[t]
\centering
\caption{CLIP Layer Selection Performance.}
\renewcommand{\arraystretch}{1.2}
\setlength{\tabcolsep}{3pt}
\resizebox{\columnwidth}{!}{%
\begin{tabular}{c|ccccccc}
\toprule[1.2pt]
\textbf{Model} & \textbf{Val.} $\downarrow$ & \textbf{FID} $\downarrow$ & \textbf{CLIP} $\uparrow$ & \textbf{Pick} $\uparrow$ & \textbf{GenE.} $\uparrow$ & \textbf{Aesth.} $\uparrow$ & \textbf{T2I.} $\uparrow$ \\
\midrule
Layer 6 & 0.428 & 15.9 & 32.3 & 20.1 & 68.4 & 5.46 & 51.2 \\
Layer 12 & 0.428 & 16.0 & 32.7 & 20.2 & 68.3 & 5.56 & 50.5 \\
Layer 18 & 0.428 & 16.1 & \textbf{32.8} & 20.2 & 69.1 & 5.56 & 51.2 \\
Layer 23 & 0.428 & 15.6 & \textbf{32.8} & \textbf{20.3} & \textbf{70.6} & \textbf{5.59} & 50.9 \\
Layer 24 & 0.428 & 16.0 & \textbf{32.8} & \textbf{20.3} & 70.4 & 5.58 & \textbf{51.4} \\
\makecell{Layer \\ 6 + 12 + 18 + 24} & 0.426 & \textbf{15.4} & \textbf{32.8} & \textbf{20.3} & 70.4 & 5.60 & 51.2 \\
\bottomrule[1.2pt]
\end{tabular}%
}
\label{tab:clip_layer_ablation}
\end{table}

\subsection{LLM}
\subsubsection{Text LLM vs.\ Multimodal LLM}

We further investigate the impact of the text encoder by comparing a text-only LLM with a multimodal LLM (MLLM). Our experiments, summarized in Table~\ref{tab:llm_vs_mllm}, reveal that text-only LLMs tend to outperform their multimodal counterparts, particularly on the GenEval metric.

\begin{table}[t]
\centering
\caption{Performance Comparison of text LLM vs.\  Multimodal LLM.}
\renewcommand{\arraystretch}{1.2}
\setlength{\tabcolsep}{3pt}
\resizebox{\columnwidth}{!}{%
\begin{tabular}{c|ccccccc}
\toprule[1.2pt]
\textbf{Model} & \textbf{Val.} $\downarrow$ & \textbf{FID} $\downarrow$ & \textbf{CLIP} $\uparrow$ & \textbf{Pick} $\uparrow$ & \textbf{GenE.} $\uparrow$ & \textbf{Aesth.} $\uparrow$ & \textbf{T2I.} $\uparrow$ \\
\midrule
\makecell{LLM}     & 0.427 & \textbf{16.0} & \textbf{32.0} & \textbf{20.1} & \textbf{72.6}  & \textbf{5.57} & 48.6 \\
\makecell{MLLM} & {0.427} & 16.4 & 31.9 & {20.0} & {70.0} & {5.54} & \textbf{49.2} \\

\bottomrule[1.2pt]
\end{tabular}%
}
\label{tab:llm_vs_mllm}
\end{table}

\subsubsection{Layer Selection in LLMs}

We also evaluated different layers in the 56-layer text-only LLM to determine the optimal choice for text embeddings. Specifically, we compared embeddings from layer 14 (early), layer 28 (middle), layer 42 (deeper), and a concatenation of layers 28 and 42. Our results, summarized in Table~\ref{tab:llm_layer_ablation}, indicate that both the middle (28) and deeper (42) layers offer a strong balance between preserving low-level token details and capturing high-level semantic representations. In contrast, the final layer, while performing well on GenEval, provides less balanced representations for text-to-image generation tasks, possibly due to over-specialization in the pretraining objective.

% \begin{table*}[t]
% \centering
% \caption{LLM Layer Selection Performance Across Key Metrics.}
% \renewcommand{\arraystretch}{1.2} % Slightly increase row height
% \setlength{\tabcolsep}{3pt} % Adjust column spacing
% \small
% \begin{tabular}{l|ccccccc}
% \toprule[1.2pt]
% \textbf{LLM Layer} & \textbf{Val Loss} $\downarrow$ & \textbf{FID} $\downarrow$ & \textbf{CLIP} $\uparrow$ & \textbf{Pick} $\uparrow$ & \textbf{Gen Eval} $\uparrow$ & \textbf{Aesth.} $\uparrow$ & \textbf{T2I Comp.} $\uparrow$ \\
% \midrule
% Layer 14       & 0.4270 & 15.79 & 32.19 & 20.07 & 68.11  & 5.562 & 0.494 \\
% Layer 28       & 0.4266 & 16.45 & 32.27 & 20.15 & 71.68  & 5.637 & 0.502 \\
% Layer 42       & 0.4269 & 17.20 & 32.24 & 20.14 & 71.60  & 5.610 & 0.501 \\
% Layer 28 + 42  & 0.4267 & 15.89 & 32.32 & 20.17 & 72.98  & 5.597 & 0.507 \\
% Layer 56 (last)& 0.4269 & 15.99 & 31.99 & 20.09 & 7.26  & 5.567 & 0.486 \\
% \bottomrule[1.2pt]
% \end{tabular}
% \label{tab:llm_layer_ablation}
% \end{table*}

\begin{table}[t]
\centering
\caption{LLM Layer Selection Performance.}
\renewcommand{\arraystretch}{1.2}
\setlength{\tabcolsep}{3pt}
\resizebox{\columnwidth}{!}{%
\begin{tabular}{c|ccccccc}
\toprule[1.2pt]
\textbf{Model} & \textbf{Val.} $\downarrow$ & \textbf{FID} $\downarrow$ & \textbf{CLIP} $\uparrow$ & \textbf{Pick} $\uparrow$ & \textbf{GenE.} $\uparrow$ & \textbf{Aesth.} $\uparrow$ & \textbf{T2I.} $\uparrow$ \\
\midrule
Layer 14       & 0.427 & \textbf{15.8} & 32.2 & 20.1 & 68.1  & 5.56 & 0.494 \\
Layer 28       & 0.427 & 16.5 & \textbf{32.3} & \textbf{20.2} & 71.7  & \textbf{5.64} & 0.502 \\
Layer 42       & 0.427 & 17.2 & \textbf{32.3} & 20.1 & 71.6  & 5.61 & 0.501 \\
Layer 56 (last)& 0.427 & 16.0 & 32.0 & 20.1 & 72.6  & 5.57 & 0.486 \\
Layer 28 + 42  & 0.427 & 15.9 & \textbf{32.3} & \textbf{20.2} & \textbf{73.0}  & 5.60 & \textbf{0.507} \\
\bottomrule[1.2pt]
\end{tabular}%
}
\label{tab:llm_layer_ablation}
\end{table}
\section{Progressive VAE Training Studies}
\label{sec:appendix vae}
To validate our progressive training approach described in Section~\ref{sec: improved vae}, we conducted experiments on three VAE variants, all employing an 8× compression factor and trained on the OpenImages 9M dataset. The variants are defined as follows:

\begin{itemize}
    \item \textbf{Variant A:} A VAE with 4 channels trained from scratch.
    \item \textbf{Variant B:} A VAE with 8 channels trained from scratch.
    \item \textbf{Variant C:} A VAE with 8 channels trained using our proposed Progressive training approach. This variant is initially trained with 4 channels (as in Variant A) and subsequently refined by replacing an intermediate convolutional layer with one that uses 8 channels.
\end{itemize}

Our evaluation employs the reconstruction FID (rFID) metric on the COCO validation set to assess image reconstruction quality, along with an evaluation of the downstream diffusion model using \model/B. The experimental results, summarized in Table~\ref{tab:vae_results}, indicate that although increasing the channel size significantly enhances reconstruction quality, it also leads to a higher KL divergence in the latent features. This elevated KL divergence can impede the latent diffusion model's learning, resulting in only marginal gains in final visual generation quality. In contrast, our progressive training pipeline mitigates this issue by first training a smaller VAE and then gradually increasing its channel capacity. This approach achieves notable improvements in text-to-image generation while maintaining competitive reconstruction performance.

\begin{table}[t]
\centering
\caption{Comparison of VAE on reconstruction and generation.}
\renewcommand{\arraystretch}{1.2} % Slightly increase row height
\setlength{\tabcolsep}{3pt} % Adjust column spacing
\resizebox{\columnwidth}{!}{%
\begin{tabular}{c|cc|ccccccc}
\toprule[1.2pt]
\textbf{Model} & \textbf{KL} &  \textbf{rFID}$\downarrow$ & \textbf{FID} $\downarrow$ & \textbf{CLIP} $\uparrow$ & \textbf{Pick} $\uparrow$ & \textbf{GenE.} $\uparrow$ & \textbf{Aesth.} $\uparrow$ & \textbf{T2I.} $\uparrow$ \\
\midrule

A & \num{7e4} & 4.62 & {17.2} & 32.7 & {20.2} & {69.8} & {5.52} & {50.4} \\

B & \num{9e5} & \textbf{2.59} & 16.3 & 32.8 & 20.2 & 69.4 & 5.56 & 50.7 \\

C & \num{7e4} & 2.61 & \textbf{16.0} & 32.8 & \textbf{20.3} & \textbf{70.4} & \textbf{5.58} & \textbf{51.4} \\
\bottomrule[1.2pt]
\end{tabular}%
}
\label{tab:vae_results}
\end{table}

\section{Supervised Fine-Tuning and Data Curation}

\label{sec:appendix sft}

In the supervised fine-tuning (SFT) stage, our goal is to refine the pretrained model using a very high-quality but relatively small dataset of image-text pairs. To this end, we curated a dataset of 1,033 pairs, ensuring that the images and their corresponding captions meet stringent quality standards.

The curation process involved several key steps:

\begin{enumerate}
    \item \textbf{Automated Filtering:} Initially, we applied both the LAION image aesthetics model~\cite{schuhmann2022laion5bopenlargescaledataset} and our internal photo aesthetics model to the pretraining data. This step allowed us to filter out images that did not meet our high aesthetic standards, ensuring that only the best images were considered.
    
    \item \textbf{Manual Selection:} From the automatically filtered subset, we manually reviewed the images to further refine the selection. The focus here was on achieving diversity across object categories and image styles, with special attention given to important verticals such as people and animals.
    
    \item \textbf{Caption Curation:} For the selected images, we crafted precise captions in the style of our automatic captioning model. This manual curation ensured that each caption was not only accurate but also semantically well-aligned with the corresponding image.
\end{enumerate}

Fine-tuning with the resulting dataset, as demonstrated in \cite{dai2023emu}, can lead the model to converge to a state where the generated images surpass the average quality of the pretraining data.

Overall, our SFT strategy emphasizes quality over quantity. By leveraging a meticulously curated dataset, we ensure that the fine-tuning process yields improved image generation performance, achieving both higher aesthetic quality and better semantic alignment.

\section{Reward Fine-tuning}
\label{sec:appendix rl}
We adopt an approach similar to DRaFT~\cite{kevin2024directlyfinetuningdiffusionmodels} to fine-tune our models using the HPSv2~\cite{xiaoshi2023humanpreferencescorev2} reward model. In our setup, the models receive a prompt \({p}\) and an initial latent noise \({\mathbf{z}_T}\) as inputs, which are then denoised over \({T}\) timesteps to generate the final image \({I_0}\). The HPSv2 model computes a human preference score for the generated image, denoted as \({r(p, I_0)}\), with scores normalized between 0 and 1. Consequently, the loss backpropagated through the sampling chain is defined as \({1 - r(p, I_0)}\).

To specifically target areas where the model underperforms, we selected 2,000 of the lowest-scoring prompts as the training data for reward fine-tuning. During this stage, we fine-tune the full set of model parameters, using a total sampling timestep \({T}=50\) and a stop gradient timestep \({T_s}=25\) to ensure that gradients are propagated only from timestep \({T}\) down to \({T_s}\), with no gradient updates before \({T_s}\).

Despite these precautions, we observed a reward model hacking phenomenon~\cite{kevin2024directlyfinetuningdiffusionmodels}, where HPSv2 occasionally assigns very high scores to poor-quality images. To mitigate this issue, we implemented an early stopping strategy to prevent overfitting and reward hacking. Ultimately, this approach effectively reduced structural artifacts, enhanced text alignment, and improved the overall visual appeal of the generated images.

\section{State-of-the-Art Model Size Breakdown}
\label{sec:appendix_sota_model_size_breakdown}

In Table~\ref{tab:sota_model_sizes}, we present a detailed breakdown of the architectural components and parameter counts for various state-of-the-art text-to-image generation models. For JanusPro-7B, the SigLIP encoder is omitted for simplicity. The comparison demonstrates that DiT-Air is relatively compact—both in total parameter count and in trainable parameters—when compared with existing models.

\begin{table*}[t]
    \centering
    \caption{Breakdown of architectural components for various text-to-image generation models. Trainable components are marked with \( \star \).}

    \label{tab:sota_model_sizes}
    \tiny
    \setlength{\tabcolsep}{4pt} % adjust column separation if needed
    \resizebox{0.9\textwidth}{!}{%
    \begin{tabular}{l l l l l r}
        \toprule
        \textbf{Model} & \textbf{Text Encoder 1} & \textbf{Text Encoder 2} & \textbf{Diffusion Model}$^\star$  & \textbf{Autoencoder} & \textbf{Total (B)} \\ \midrule
        SDXL Base          & CLIP/L (123M)   & OpenCLIP/g (694M) & U-Net (2.6B) & 8-ch (84M)  & 3.50 \\
        PixArt-$\alpha$    & Flan-T5-XXL (4.7B)  & --                & DiT (0.6B)  & 4-ch (80M)  & 5.38 \\
        SD3 Medium         & CLIP/L + bigG (817M) & T5-v1.1-XXL (4.7B) & DiT (2.0B)  & 16-ch (85M) & 7.65 \\
        SD3                & CLIP/L + bigG (817M) & T5-v1.1-XXL (4.7B) & DiT (8.0B)  & 16-ch (85M) & 13.60 \\
        Flux-Dev           & CLIP/L (123M)   & T5-XXL (4.7B)     & DiT (12B)   & 16-ch (85M) & 16.91 \\
        Flux-Schnell       & CLIP/L (123M)   & T5-XXL (4.7B)     & DiT (12B)   & 16-ch (85M) & 16.91 \\
        JanusPro-7B      & --                  & --                & LLM (6.9B)         & 16-ch VQ (85M) & 6.91 \\ 
        \midrule
        \model/L-Lite &CLIP/H (335M) & -- & DiT (0.7B) & 8-ch (84M) & 1.15\\
        \model/XXL &CLIP/H (335M) & LLM (2.8B) & DiT (2.8B) & 8-ch (84M) & 5.95\\
        \bottomrule
    \end{tabular}%
    }
\end{table*}

%#\clearpage % Start a new page
\section{Generation Examples} % Use * to avoid numbering
\begin{figure*}[t] % Keep it within the same page
    \centering
    \includegraphics[height=1\textheight,trim=0 20 0 20, clip, keepaspectratio]{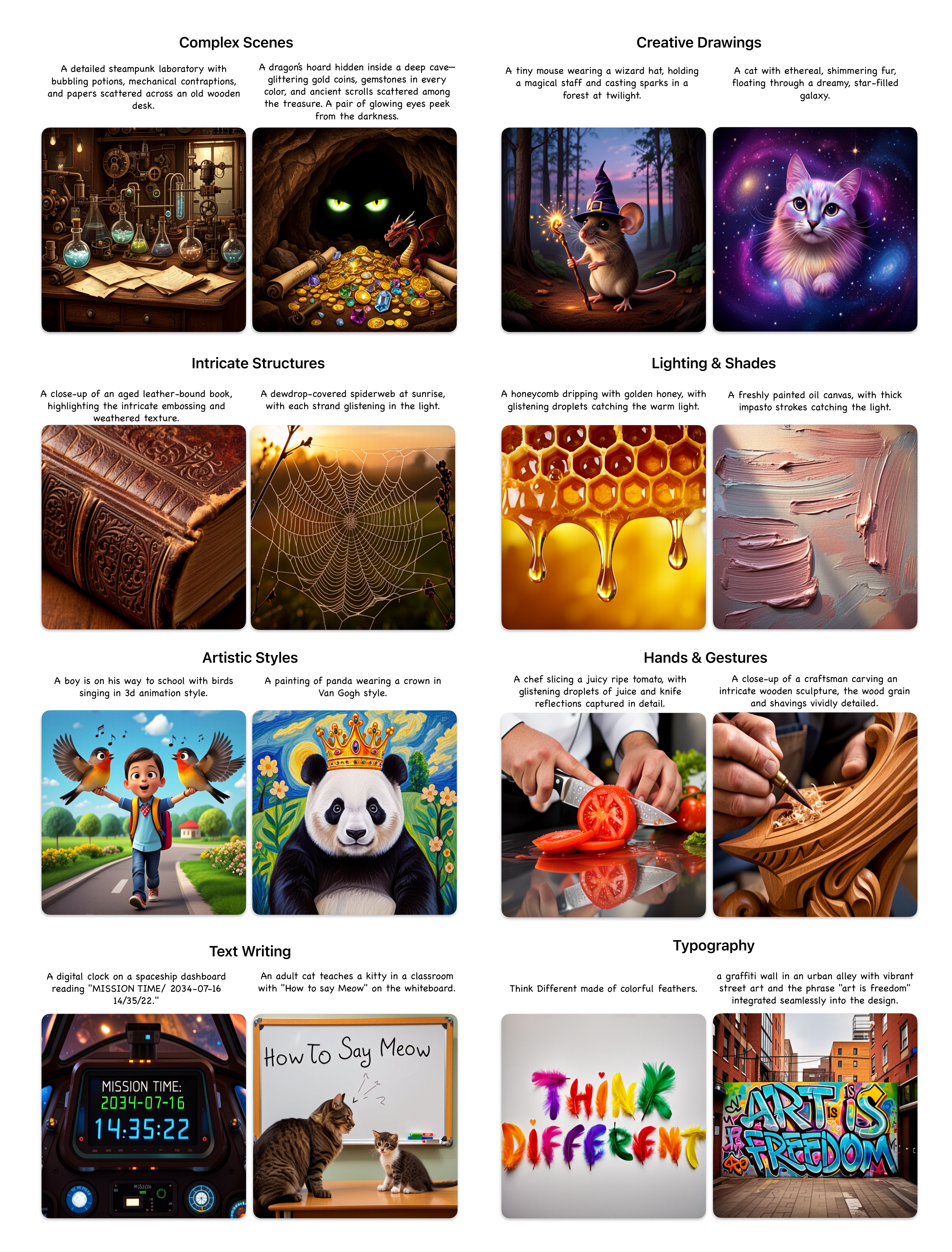}
    \caption{Sample images from our \model/XXL illustrating different capabilities. }
    \label{fig:sample_pdf}
\end{figure*}

\begin{figure*}[t] % Keep it within the same page
    \centering
    \includegraphics[height=1\textheight,trim=0 20 0 20, clip, keepaspectratio]{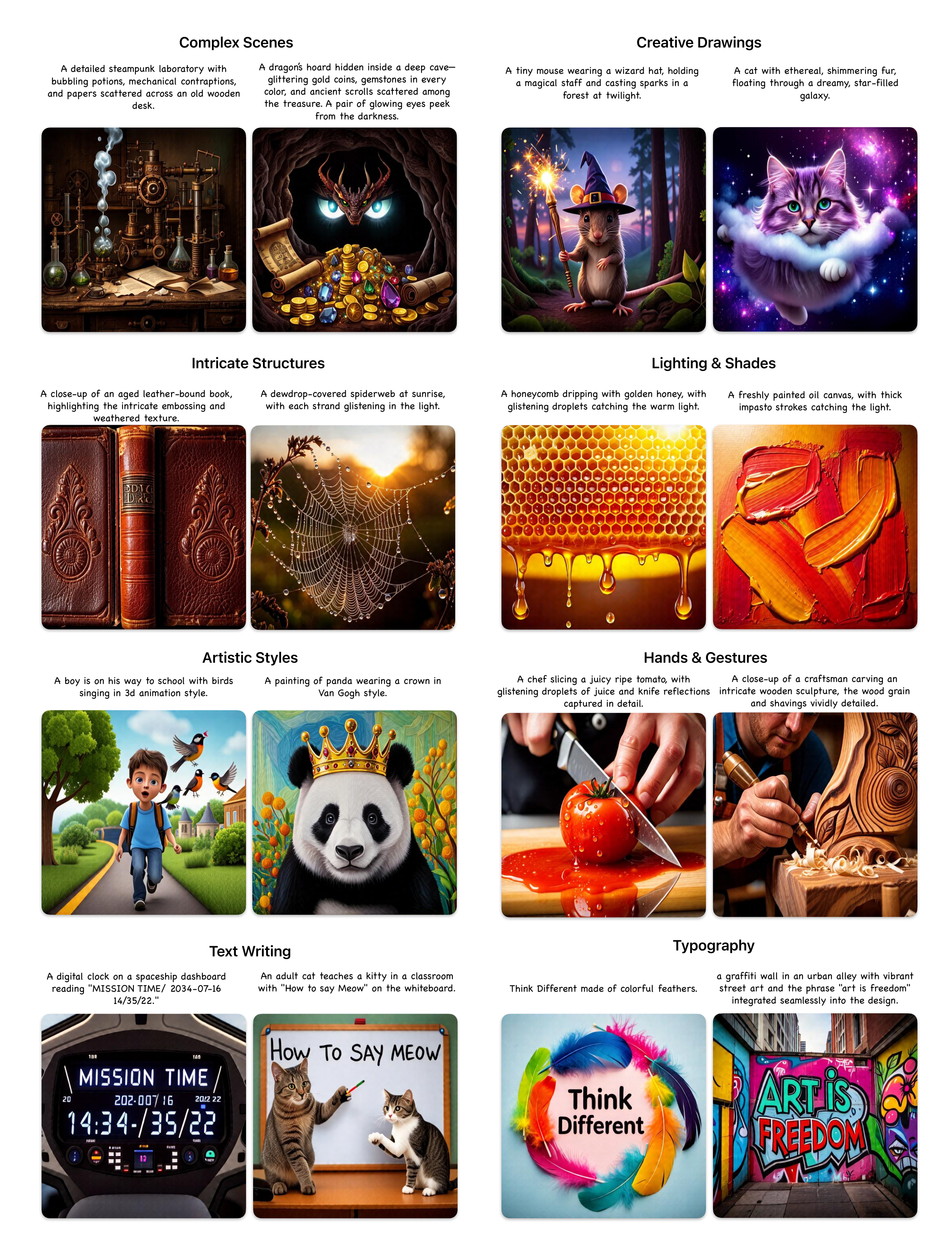}
    \caption{Sample images from our \model/L-Lite illustrating different capabilities. }
    \label{fig:lite_sample_pdf}
\end{figure*}

\label{sec:appendix_samples}
A selection of model-generated images illustrating different capabilities are shown in Figure~\ref{fig:sample_pdf} (generations by~\model/XXL) and Figure~\ref{fig:lite_sample_pdf} (generations by~\model/L-Lite).

Through qualitative observation, we find that the ~\model/XXL consistently excels in generating complex scenes, capturing intricate structural details, and delivering superior visual quality. It demonstrates a strong ability to produce highly detailed and realistic images, even when handling lengthy or complex prompts. Its integrated LLM encoder contributes significantly to its robust text rendering capabilities, maintaining clarity and accuracy even in challenging scenarios.

In contrast, the~\model/L-Lite offers a well-balanced approach, prioritizing efficiency while still delivering strong performance across a variety of tasks. It is particularly well-suited for scenarios with limited computational resources, providing high-quality images and effective handling of most prompts. While the \model/L-Lite maintains an excellent balance of efficiency and quality, our qualitative observations indicate that it may occasionally struggle with more ambiguous prompts and exhibit limitations in rendering complex text or achieving the same level of visual fidelity as the \model/XXL. These observations highlight the intended trade-off of the \model/L-Lite: a streamlined design that prioritizes efficiency, making it a compelling choice when balancing performance with resource constraints.

Overall, these observations highlight the trade-offs between the two models. The \model/XXL is ideal for tasks that demand high-quality, detailed images and strong text-image alignment, whereas the \model/L-Lite serves as a compelling, resource-efficient alternative for more lightweight use cases.

\vfill % Push any remaining content down to balance the page
%\clearpage % Ensure the next content starts on a new page
\begin{table*}[ht]
\centering
\caption{Supplementary Table: Detailed GenEval and T2I CompBench Breakdown.}
\resizebox{\textwidth}{!}{%
\begin{tabular}{l|ccccccc|ccccccc}
\toprule[1.2pt]
\textbf{Method} & \multicolumn{7}{c|}{\textbf{GenEval}} & \multicolumn{7}{c}{\textbf{T2I Compbench} (\%)} \\
\midrule
                & \textbf{Overall} & \textbf{Single Object} & \textbf{Two Objects} & \textbf{Counting} & \textbf{Colors} & \textbf{Position} & \textbf{Color Attribution} & \textbf{Average} & \textbf{Color} & \textbf{Shape} & \textbf{Texture} & \textbf{Spatial} & \textbf{Non-Spatial} & \textbf{Complex} \\
\midrule
\multicolumn{15}{c}{\textbf{Table~\ref{tab:mmdits_comparison}}} \\
\midrule
Per-layer Adaln & 69.8 & 99.3 & 88.7 & 68.6 & 81.0 & 22.0 & 59.3 & 51.1 & 81.8 & 62.2 & 74.9 & 19.6 & 31.3 & 37.1 \\
Shared Adaln    & 69.8 & 99.3 & 90.0 & 68.2 & 79.1 & 23.5 & 58.7 & 51.0 & 81.8 & 61.9 & 73.3 & 19.9 & 31.3 & 37.7 \\
\midrule
\multicolumn{15}{c}{\textbf{Table~\ref{tab:monodit_lite_comparison}}} \\
\midrule
\model/B                     & 70.4 & 99.7 & 88.6 & 68.1 & 80.0 & 25.2 & 60.8 & 51.4 & 83.0 & 61.7 & 73.4 & 21.6 & 31.3 & 37.7 \\
\model/B-lite (full)         & 58.4 & 96.8 & 69.3 & 49.8 & 75.7 & 14.1 & 44.4 & 47.6 & 79.1 & 57.4 & 69.4 & 13.5 & 30.8 & 35.6 \\
\model/B-lite (attention)    & 66.9 & 99.1 & 84.3 & 65.2 & 79.2 & 18.5 & 54.9 & 49.6 & 79.9 & 60.3 & 71.2 & 18.0 & 31.2 & 37.1 \\
\midrule
\multicolumn{15}{c}{\textbf{Figure~\ref{fig:architecture_comparison}}} \\
\midrule
Pixart-$\alpha$/S   & 61.9 & 98.7 & 81.2 & 58.3 & 79.4 & 15.3 & 52.0 & 48.3 & 80.2 & 57.5 & 68.8 & 15.8 & 31.0 & 36.2 \\
Pixart-$\alpha$/B   & 68.1 & 99.5 & 89.4 & 67.3 & 79.6 & 21.2 & 57.7 & 50.5 & 82.7 & 60.5 & 72.5 & 18.9 & 31.4 & 37.1 \\
Pixart-$\alpha$/L   & 70.4 & 99.2 & 89.1 & 69.5 & 80.4 & 24.2 & 58.2 & 50.9 & 82.7 & 61.0 & 73.0 & 19.5 & 31.4 & 37.7 \\
Pixart-$\alpha$/XL  & 70.1 & 99.2 & 88.6 & 68.2 & 83.4 & 26.1 & 57.9 & 50.9 & 81.9 & 61.3 & 72.7 & 20.4 & 31.3 & 37.5 \\
Pixart-$\alpha$/XXL & 69.9 & 99.2 & 88.6 & 71.2 & 81.9 & 26.8 & 56.5 & 51.0 & 82.3 & 63.0 & 72.8 & 19.0 & 31.2 & 37.8 \\
\midrule
MMDiT/S   & 67.0 & 99.5 & 89.4 & 73.6 & 83.1 & 29.1 & 58.8 & 50.0 & 80.6 & 60.8 & 72.7 & 18.2 & 31.0 & 36.5 \\
MMDiT/B   & 69.8 & 99.3 & 88.7 & 68.6 & 81.0 & 22.0 & 59.3 & 51.1 & 81.8 & 62.2 & 74.9 & 19.6 & 31.3 & 37.1 \\
MMDiT/L   & 70.9 & 99.3 & 90.9 & 71.2 & 82.3 & 23.8 & 58.3 & 51.1 & 82.3 & 62.2 & 74.4 & 19.1 & 31.3 & 37.2 \\
MMDiT/XL  & 69.0 & 98.9 & 87.3 & 66.3 & 81.1 & 26.4 & 53.9 & 50.9 & 82.6 & 63.0 & 73.1 & 18.1 & 31.2 & 37.4 \\
MMDiT/XXL & 69.7 & 99.3 & 87.4 & 67.1 & 83.0 & 24.3 & 56.9 & 50.4 & 82.3 & 62.7 & 73.2 & 15.7 & 31.3 & 37.2 \\
\midrule
\model/S   & 63.4 & 98.0 & 78.2 & 57.2 & 79.0 & 15.0 & 53.1 & 48.9 & 80.1 & 59.1 & 71.6 & 15.2 & 31.0 & 36.6 \\
\model/B   & 70.4 & 99.7 & 88.6 & 68.1 & 80.0 & 25.2 & 60.8 & 51.4 & 83.0 & 61.7 & 73.4 & 21.6 & 31.3 & 37.7 \\
\model/L   & 69.6 & 99.4 & 88.4 & 67.1 & 79.5 & 24.7 & 58.9 & 51.2 & 82.6 & 62.5 & 73.7 & 19.6 & 31.1 & 37.5 \\
\model/XL  & 71.1 & 99.3 & 90.0 & 71.4 & 82.1 & 26.8 & 56.8 & 51.1 & 82.6 & 62.0 & 73.4 & 19.9 & 31.2 & 37.6 \\
\model/XXL & 69.7 & 99.0 & 89.0 & 70.1 & 81.3 & 24.9 & 54.3 & 50.8 & 82.7 & 62.6 & 73.2 & 17.9 & 31.5 & 36.8 \\
\midrule
\multicolumn{15}{c}{\textbf{Table~\ref{tab:text_encoder_summary}}} \\
\midrule
CLIP (Bidirectional) & 70.4 & 99.3 & 88.7 & 68.6 & 81.0 & 22.0 & 59.3 & 51.1 & 81.8 & 62.2 & 74.9 & 19.6 & 31.3 & 37.1 \\
LLM (Text-only)      & 72.6 & 99.2 & 88.7 & 65.7 & 83.6 & 34.7 & 63.6 & 48.6 & 78.8 & 56.8 & 69.8 & 18.4 & 30.9 & 37.1 \\
T5-XXL               & 66.9 & 65.3 & 82.7 & 63.9 & 78.3 & 19.5 & 49.6 & 48.0 & 76.3 & 57.5 & 69.1 & 17.6 & 31.1 & 36.5 \\
\midrule
\multicolumn{15}{c}{\textbf{Table~\ref{tab:sota_comparison}}} \\
\midrule

SDXL        & 55.7 & 98.0 & 74.0 & 39.0 & 85.0 & 15.0 & 23.0 
            & 40.6 & 58.8 & 46.9 & 53.0 & 21.3 & 31.2 & 32.4 \\

PixArt-$\alpha$ 
            & 47.8 & 98.0 & 50.0 & 44.0 & 80.0 & 8.00 & 7.00 
            & 44.6 & 66.9 & 49.3 & 64.8 & 20.6 & 32.0 & 34.3 \\

SD3-medium  & 62.0 & 98.0 & 74.0 & 63.0 & 67.0 & 34.0 & 36.0 
            & 52.4 & 81.3 & 58.9 & 73.3 & 32.0 & 31.4 & 37.7 \\

SD3         & 74.5 & 99.0 & 94.0 & 72.0 & 89.0 & 33.0 & 60.0 
            & 51.4 & --   & --   & --   & --   & --   & --   \\

Flux-dev    & 66.7 & 99.0 & 81.0 & 79.0 & 74.0 & 20.0 & 47.0 
            & 49.6 & 74.1 & 57.2 & 69.2 & 28.6 & 31.3 & 37.0 \\

Flux-schnell
            & 70.7 & 99.0 & 92.0 & 73.0 & 78.0 & 28.0 & 54.0 
            & 49.9 & 73.9 & 55.8 & 68.5 & 31.2 & 31.5 & 38.6 \\

Janus-pro   & 80.3 & 99.0 & 89.0 & 59.0 & 90.0 & 79.0 & 66.0 
            & 35.2 & 52.0 & 33.1 & 40.6 & 15.4 & 31.3 & 39.2 \\

\model/L-Lite & 77.6 & 99.9 & 96.5 & 80.8 & 86.8 & 30.5 & 71.3 & 55.4 & 86.1 & 66.6 & 79.5 & 27.7 & 31.2 & 41.3 \\
\model/XXL  & 82.9 & 99.9	& 98.8 & 83.8 & 86.4 & 50.5 &	77.7
 & 59.5	& 88.9 & 69.4 & 82.0 & 40.6 &  31.9 & 44.1
 \\
\midrule
\multicolumn{15}{c}{\textbf{Table~\ref{tab:multi_stage_training}}} \\
\midrule
Pretrain 256\(^2\)  & 71.1 & 99.3 & 88.7 & 69.0 & 82.0 & 32.0 & 55.7 & 50.7 & 81.9 & 61.2 & 72.4 & 19.5 & 31.3 & 37.9 \\
Pretrain 512\(^2\)  & 74.2 & 99.4 & 91.7 & 73.6 & 84.0 & 35.4 & 61.4 & 51.7 & 81.9 & 61.2 & 73.9 & 22.6 & 31.1 & 39.6 \\
Supervised fine-tuning  & 79.0 & 99.9 & 96.3 & 78.7 & 86.1 & 44.6 & 68.5 & 55.3 & 85.5 & 63.4 & 76.9 & 32.7 & 31.6 & 41.9 \\
Reward fine-tuning & 82.9 & 100.0 & 98.8 & 83.8 & 86.6 & 50.5 & 77.7 & 59.5 & 88.9 & 69.4 & 82.0 & 40.6 & 31.9 & 44.2 \\
\midrule
\multicolumn{15}{c}{\textbf{Table~\ref{tab:causal_vs_bidirectional_clip}}} \\
\midrule
Bidirectional & 70.4 & 99.7 & 88.6 & 68.1 & 80.0 & 25.2 & 60.8 & 51.4 & 83.0 & 61.7 & 73.4 & 21.6 & 31.3 & 37.7 \\
Causal        & 68.3 & 99.5 & 83.9 & 67.3 & 79.0 & 21.9 & 58.5 & 50.60 & 81.7 & 61.5 & 72.3 & 19.3 & 31.4 & 37.2 \\
\midrule
\multicolumn{15}{c}{\textbf{Table~\ref{tab:clip_layer_ablation}}} \\
\midrule
Layer 6                & 68.4 & 99.0 & 85.6 & 64.9 & 79.2 & 23.6 & 58.2 & 51.2 & 81.2 & 61.6 & 74.0 & 22.4 & 31.1 & 37.2 \\
Layer 12               & 68.3 & 99.5 & 87.0 & 66.0 & 80.0 & 20.9 & 56.7 & 50.5 & 81.7 & 61.0 & 73.2 & 18.6 & 31.2 & 37.5 \\
Layer 18               & 69.1 & 99.2 & 89.1 & 66.7 & 79.5 & 22.1 & 58.2 & 51.2 & 82.4 & 61.5 & 74.3 & 20.5 & 31.0 & 37.4 \\
Layer 23               & 70.6 & 99.3 & 88.9 & 69.7 & 80.9 & 24.1 & 60.8 & 50.9 & 82.1 & 61.1 & 73.7 & 19.5 & 31.3 & 37.7 \\
Layer 24               & 70.4 & 99.7 & 88.6 & 68.1 & 80.0 & 25.2 & 60.8 & 51.4 & 83.0 & 61.7 & 73.4 & 21.6 & 31.3 & 37.7 \\
Layer 6 + 12 + 8 + 24  & 70.4 & 99.6 & 91.0 & 67.9 & 79.6 & 26.1 & 58.5 & 51.2 & 81.8 & 61.3 & 74.1 & 21.0 & 31.2 & 37.5 \\
\midrule
\multicolumn{15}{c}{\textbf{Table~\ref{tab:llm_vs_mllm}}} \\
\midrule
LLM   & 70.2 & 99.4 & 87.9 & 67.4 & 81.2 & 24.9 & 60.1 & 51.4 & 83.0 & 61.0 & 74.2 & 21.5 & 31.2 & 37.8 \\
MLLM  & 69.6 & 99.5 & 88.5 & 67.1 & 80.9 & 25.2 & 56.6 & 51.3 & 82.0 & 62.2 & 73.6 & 20.7 & 31.4 & 37.8 \\
\midrule
\multicolumn{15}{c}{\textbf{Table~\ref{tab:llm_layer_ablation}}} \\
\midrule
Layer 14        & 68.1 & 98.9 & 82.0 & 59.2 & 80.7 & 27.3 & 60.6 & 49.4 & 80.4 & 59.3 & 68.7 & 19.8 & 31.3 & 37.1 \\
Layer 28        & 71.7 & 99.0 & 87.6 & 62.6 & 82.6 & 34.3 & 64.0 & 50.2 & 81.1 & 60.0 & 72.0 & 19.9 & 31.1 & 37.1 \\
Layer 42        & 71.6 & 98.4 & 86.1 & 63.3 & 82.8 & 37.1 & 61.8 & 50.1 & 79.5 & 60.0 & 72.0 & 20.7 & 31.0 & 37.6 \\
Layer 56 (last) & 70.2 & 99.4 & 88.0 & 67.4 & 81.2 & 24.9 & 60.1 & 51.4 & 83.0 & 61.0 & 74.2 & 21.5 & 31.2 & 37.8 \\
Layer 28 + 42   & 73.0 & 99.2 & 87.7 & 66.3 & 85.0 & 33.0 & 66.6 & 50.7 & 81.1 & 61.1 & 73.0 & 20.3 & 31.1 & 37.7 \\
\midrule
\multicolumn{15}{c}{\textbf{Table~\ref{tab:vae_results}}} \\
\midrule
A & 69.8 & 99.2 & 88.5 & 69.9 & 79.1 & 25.7 & 55.9 & 50.4 & 80.3 & 61.8 & 72.9 & 19.3 & 31.3 & 37.0 \\
B & 69.4 & 99.6 & 89.9 & 67.4 & 79.8 & 23.0 & 56.9 & 50.7 & 81.1 & 61.0 & 73.1 & 21.2 & 31.1 & 36.5 \\
C & 70.4 & 99.7 & 88.6 & 68.1 & 80.0 & 25.2 & 60.8 & 51.4 & 83.0 & 61.7 & 73.4 & 21.6 & 31.3 & 37.7 \\

\bottomrule[1.2pt]
\end{tabular}%
}
\label{tab:combined_detailed_metrics}
\end{table*}

\end{document}